\documentclass{article}

\PassOptionsToPackage{numbers, compress}{natbib}
\usepackage[main, final]{neurips_2026}

\usepackage[utf8]{inputenc} % allow utf-8 input
\usepackage[T1]{fontenc}    % use 8-bit T1 fonts
\usepackage{hyperref}       % hyperlinks
\usepackage{url}            % simple URL typesetting
\usepackage{booktabs}       % professional-quality tables
\usepackage{amsfonts}       % blackboard math symbols
\usepackage{nicefrac}       % compact symbols for 1/2, etc.
\usepackage{microtype}      % microtypography
\usepackage{xcolor}         % colors
\usepackage{fontawesome5}
\usepackage{multicol}
\usepackage{multirow}
\usepackage{adjustbox}
\usepackage{makecell}     
\usepackage{pifont}       
\usepackage{caption}      
\usepackage{array}
\usepackage{amsmath}
\usepackage{algorithm}
\usepackage{algpseudocode}
\usepackage{amssymb}
\usepackage{tikz}
\usetikzlibrary{arrows.meta,positioning,shapes.geometric}
\usepackage{wrapfig}
\usepackage[most]{tcolorbox}
\usepackage{listings}
\usepackage{xcolor}
\definecolor{boxtitle}{RGB}{230,230,230}
\definecolor{boxframe}{RGB}{180,180,180}
\definecolor{boxback}{RGB}{248,248,248}

\usepackage[table]{xcolor} 
\usepackage{colortbl}

\definecolor{myyellow}{RGB}{255,165,0} 
\definecolor{myblue}{RGB}{0,0,255}

\newcommand{\correctspan}[1]{\begingroup\setlength{\fboxsep}{1.2pt}\colorbox{green!12}{#1}\endgroup}
\newcommand{\errorspan}[1]{\begingroup\setlength{\fboxsep}{1.2pt}\colorbox{red!12}{#1}\endgroup}

\algrenewcommand{\algorithmiccomment}[1]{\hfill{\color{gray}$\triangleright$~#1}}

\newcommand{\method}{CopyGen}

\title{When Copying Is Hard: Copy-Constrained Decoding for Exact Span Reproduction}

\author{%
  Jinghui Zhang$^{1}$ \quad
  Lang Gao$^{1}$\quad
  Zongfang Liu$^{2,3}$ \quad
  Ruihong Zeng$^{1}$ \\
  \bfseries Zirui Song$^{1}$ \quad
  Rui Yan$^{4,5}$ \quad
  Kentaro Inui$^{1,6,7}$ \quad
  Xiuying Chen$^{1}$\thanks{Corresponding author.} \\
  $^{1}$MBZUAI  \quad
  $^{2}$Zhejiang University  \quad
  $^{3}$Westlake University \\
  $^{4}$Wuhan University  \quad
  $^{5}$Wuhan Textile University\quad
  $^{6}$Tohoku University  \quad
  $^{7}$RIKEN \\
  \texttt{\{jinghui.zhang, kentaro.inui, xiuying.chen\}@mbzuai.ac.ae}
}

\begin{document}

\maketitle

\begin{abstract}
Precisely reproducing a target span verbatim within a long context is a fundamental capability test for LLMs: even when the target string is present in context, models often still drift within the span or fail to stop correctly, revealing limitations in source grounding and boundary control.
This capability has broad practical relevance. For instance, code generation often requires embedding sensitive strings such as API keys exactly as they appear in context.
In this work, we construct a benchmark to test exact span reproduction ability across contracts, scientific literature, code, and random sequences, and find that current models still make surprisingly frequent errors.
To address this, we propose \textbf{\method}, a lightweight copy-constrained plugin for frozen LLMs that turns copying from a byproduct of next-token prediction into a controlled process.
Our key idea is to decompose copying into three distinct decisions: when to copy, how far to copy, and when to stop. 
Concretely, \method{} first determines whether to enter copy mode, then predicts a tentative copy length, and finally verifies the generated span token by token. 
During verification, we adaptively adjust the logits of copy continuation tokens to incentivize or suppress continuation, and emit only the longest valid prefix.
Across three LLM backbones and four benchmark tasks, \method{} improves average exact match by 35\% relative without model finetuning, and integrates seamlessly with SFT to further improve it by 4\%, while substantially accelerating generation.
These results suggest that reliable exact copying is not something LLMs obtain for free, but a decoding-time control problem that requires explicit modeling.
Code is available at \texttt{\url{https://github.com/mbzuai-nlp/CopyGen}}.
\end{abstract}

\section{Introduction}

Precisely reproducing long spans of text verbatim, referred to as instruction-conditioned long-span exact copying, can be used to evaluate large language models’ ability in exact copying and controlled generation.
Although it appears simple, this task places nontrivial demands on models’ capabilities in sequence alignment, boundary identification, and fine-grained control over generation. At the same time, this capability has important practical value. 
In software engineering workflows, models must preserve API keys, identifiers, or configuration strings exactly as they appear in context~\citep{jimenez2024swebench}. In legal processing, they must reproduce clauses verbatim rather than paraphrase them~\citep{wang2025acord}.
In such settings, a single incorrect character can break execution or alter the meaning of a legally sensitive passage. Furthermore, when the desired span is already present in the input, regenerating it token by token is both error-prone and computationally unnecessary, making exact copying not only a practically important capability but also an opportunity for improved efficiency.

Despite its practical importance, long-span exact copying has not been studied as a capability in its own right.
Existing work on copying falls into two lines, neither of which addresses this problem directly. 
Architectural copy mechanisms such as CopyNet \citep{gu-etal-2016-incorporating} and pointer-generator networks \citep{see-etal-2017-get} help models reuse source tokens to improve generation quality, but do not target character-level exactness. 
% Efficiency-oriented methods such as LLMA \citep{yang2023inference} and CopySpec \citep{dumitru-etal-2025-copyspec} exploit copying to accelerate inference, with correctness verified against the base model's own predictions rather than strengthened by the copy process itself.
Efficiency-oriented methods such as speculative decoding~\citep{leviathan2023fast}, LLMA~\citep{yang2023inference}, and CopySpec~\citep{dumitru-etal-2025-copyspec} accelerate inference through multi-token proposal or reuse, with correctness verified against the base model's own predictions rather than strengthened by the copy process itself.

To make this problem concrete, we construct a benchmark for instruction-conditioned exact span reproduction that focuses on preserving the required source text verbatim during generation, rather than broader questions of semantic faithfulness. 
Specifically, the benchmark covers three real-world tasks, namely legal contracts, scientific literature, and code contexts, where even minor deviations are unacceptable. 
We further include a synthetic random-sequence task to stress-test copying performance under extreme conditions.
Using this benchmark, we evaluate a range of LLMs and find that standard decoding fails consistently across model families and sizes.
The errors mainly take two characteristic forms.
Internal token mismatches arise when the model drifts mid-span, substituting a plausible-but-wrong token for the correct one. 
Boundary overruns occur when the model continues generating along the source span even after the target span should have ended.
These failures suggest that exact copying requires stronger control than standard next-token prediction provides, calling for an explicit copying mechanism that can maintain source alignment and enforce the correct stopping boundary.

To tackle this issue, we propose \textbf{\method}, a lightweight copy-constrained plugin for frozen LLMs. 
The key idea is to treat exact copying as a controlled decoding problem rather than a byproduct of next-token prediction. 
\method{} decomposes exact copying into three coupled decisions: when to copy, how far to tentatively copy, and when to stop. 
Concretely, at each decoding step, \method{} measures how strongly the current output prefix is anchored to the source. Once sufficient grounding is detected, the plugin proposes a tentative copy span and replays it token by token against the frozen model. 
Our analysis reveals tokens that should be copied are predicted with a large logits margin over competing alternatives, while errors, whether mid-span drifts or boundary overruns, concentrate precisely where that margin narrows. 
During verification, we utilize a learned bias to calibrate the margin at each replayed position, reinforcing source-following where confidence is high and suppressing it where the boundary is near, emitting only the longest prefix that clears the threshold.

Across three LLM backbones and four benchmark tasks, \method{} delivers substantial improvements in both copy accuracy and decoding efficiency. When applied to frozen models, it improves average exact match by approximately 35\% relative. When combined with supervised fine-tuning (SFT), CopyGen continues to provide gains, further increasing average exact match by about 4\% relative while also significantly improving decoding throughput. These results suggest that reliable exact copying is not an emergent byproduct of LLM training, but rather a capability that requires explicit control at decoding time along with dedicated copying mechanisms.

In summary, our contributions are threefold:
(1) We formulate instruction-conditioned long-span exact copying as a distinct evaluation problem and introduce a dedicated benchmark for studying it.
(2) We propose \method{}, a lightweight copy-constrained plugin for frozen LLMs that augments frozen LLMs with explicit copy control during decoding.
(3) We demonstrate that \method{} substantially improves exact-copy reliability and decoding throughput across model families and scales, and remains effective when combined with supervised finetuning.

\section{Related Work}

\textbf{Grounded and Faithful Generation.}
A closely related line of work studies how to make model outputs stay faithful to an external source rather than drifting toward unsupported continuations. 
This includes constrained generation frameworks for truthful dialogue grounded in symbolic representations \citep{fang-etal-2023-whole}, methods for verifiable generation with fine-grained citations to supporting documents \citep{cao-wang-2024-verifiable}, and approaches that improve context adherence by adapting the model or prompt at inference time \citep{zhu-etal-2024-fastmem}.
In retrieval-augmented settings, faithfulness has also been studied as the problem of making generation follow retrieved evidence, especially when retrieved context conflicts with the model's parametric knowledge \citep{zhang-etal-2025-faithfulrag}.
These works constrain generation to respect an external source, but target semantic faithfulness rather than character-level exactness. Our setting is stricter: even a single character error or formatting corruption already constitutes failure.

\textbf{Constrained and Controlled Decoding.}
Constrained Decoding aims to control model behavior directly at inference time instead of modifying the full model.
Early methods were primarily designed to enforce formal validity, for example by restricting generation to syntactically admissible outputs, as in PICARD \citep{scholak-etal-2021-picard} for text-to-SQL decoding.
More recent approaches have extended this idea beyond syntax to broader inference-time control, including symbolic or grammar-based constraints in grounded generation \citep{fang-etal-2023-whole,park2024grammaraligned}, relation-level constraints \citep{chen2022relationconstrained}, energy-based constrained generation \citep{qin2022cold}, and efficient non-invasive constraint enforcement \citep{pmlr-v235-beurer-kellner24a}.
\method{} is similarly a decoding-time intervention, but differs in the type of constraint it imposes. 
It is a source-grounded copy constraint that evolves online with the generated prefix, requiring decisions not only on whether to copy, but also how far to extend and where to stop.

\textbf{Copying and Exact Reproduction in Language Models.}
Copy mechanism has a long history in sequence generation. 
Early neural models introduced explicit copy distributions or pointer mechanisms to reproduce source tokens \citep{gu-etal-2016-incorporating,see-etal-2017-get}, and later extended them to more explicit span-level copying and alignment \citep{panthaplackel2021copy}.
More recent inference-time methods also reuse source spans during decoding, primarily to accelerate generation \citep{yang2023inference,dumitru-etal-2025-copyspec} or improve generation quality \citep{li2024nearest,liu-etal-2024-generation}, rather than to ensure exact reproduction itself. 
A separate thread studies copying as emergent or potentially risky behavior, including controlling memorized content extraction \citep{ozdayi-etal-2023-controlling}, studying probabilistic extraction of memorized content \citep{hayes-etal-2025-measuring}, and analyzing copying as a pretraining ability \citep{lv-etal-2025-language}. 
Our work differs from all of these: we focus on instruction-conditioned exact reproduction of spans explicitly present in the current context, and pair this with a dedicated benchmark and a decoding mechanism designed to improve exact-copy reliability.

%In particular, these methods verify copied or speculative continuations against the base model, so correctness remains tied to the model's original token preferences rather than being strengthened by the copy process itself. 

\vspace{-2mm}

\section{Method}
\label{sec:method}
\vspace{-1mm}

% \subsection{Problem Formulation}

We consider instruction-conditioned exact copying from long source documents. Each example is a triple $(x_{1:M}, r, y_{1:T})$, where $x_{1:M}$ is a source document, $r$ is a user instruction, and $y_{1:T}$ is the target output, all represented in the token space of a frozen base LM with vocabulary $\mathcal{V}$.
The target $y_{1:T}$ consists of two interleaved components: spans copied verbatim from the source, and novel tokens not present in the source. 
Our goal is to recover $y_{1:T}$ exactly: for copied spans, every character, boundary, and formatting detail must be preserved; for novel tokens, generation should proceed faithfully as usual. 
Our preliminary experiments show that standard autoregressive decoding is brittle for copy-heavy outputs: small local mistakes can truncate a copied span, omit formatting, or continue beyond the correct boundary. 
We therefore propose \method, a lightweight copy-constrained method for frozen LLMs that governs three decisions: when to start copying (\S\ref{trigger}), how far to propose copying (\S\ref{span}), and how much of the proposal to verify and emit (\S\ref{verification}).

\begin{figure}[t]
  \centering
  \includegraphics[width=\linewidth]{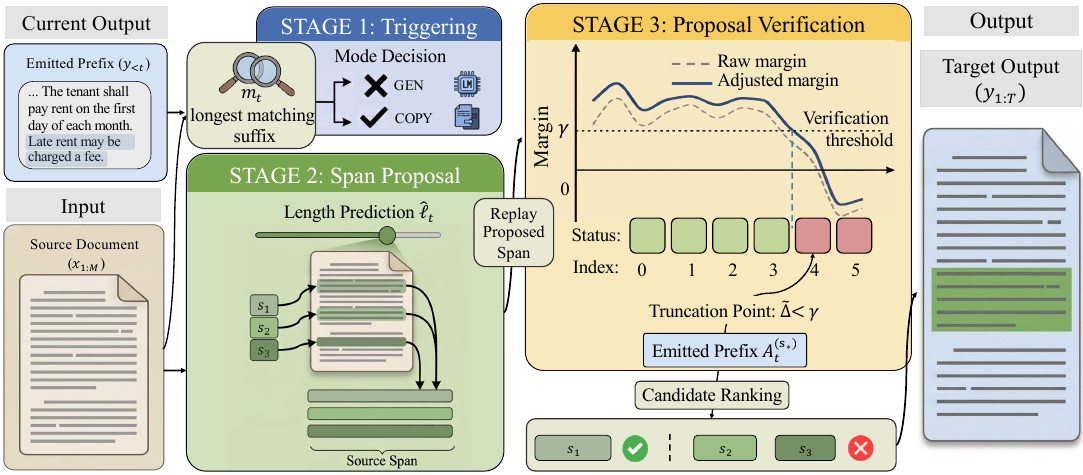}
  \caption{
  \textbf{Overview of \method{} architecture.} The system operates in three stages: (1) Copy Triggering: determining when to enter copy mode based on grounding length $m_t$ and a learned trigger head; (2) Span Proposal: predicting an adaptive copy length $\hat{\ell}_t$ in log space; and (3) Proposal Verification: calibrating token margins to decide the final emitted prefix $A_t^{(s)}$. This design supports exact copying while maintaining the flexibility of the base LM.}
  \label{fig:method_overview}
  \vspace{-5mm}
\end{figure}

\vspace{-2mm}
\subsection{Copy Triggering}
\label{trigger}

Reliable copying requires knowing precisely when to start. Entering copy mode too early risks producing spurious repetitions, while entering too late causes the model to miss the correct span entirely. We therefore design a two-stage trigger: first measuring how strongly the current prefix is anchored to the source, then using a learned head to make the final decision.

At decoding step $t$, let $y_{<t}$ denote the emitted prefix. We define the grounding length
\begin{align}
    m_t = \max\{\ell \in \{0,\dots,\min(t-1,M)\} : \exists j \in \{\ell,\dots,M\},\; x_{j-\ell+1:j} = y_{t-\ell:t-1}\},
\end{align}
where $m_t$ is the length of the longest suffix of the current output prefix that appears contiguously in the source.
We also collect the continuation positions compatible with this longest grounded suffix:
\begin{align}
    \mathcal{C}_t = \{j+1 : x_{j-m_t+1:j} = y_{t-m_t:t-1},\; j < M\},
\end{align}
Each element of $\mathcal{C}_t$ points to a source position that can plausibly continue the current prefix. The scalar $m_t$ provides a compact measure of source grounding, ensuring that copy mode is not activated by accidental overlap on a common token or punctuation mark. If $m_t$ is below a threshold $\tau$, the decoder remains in ordinary generation mode; otherwise the grounding state is passed to \method\ together with the frozen model hidden state.
 
Even when the prefix is strongly anchored to the source, the model may already be near the end of the relevant span, or the remaining continuation may no longer belong to the desired target.
We therefore introduce a trigger head that jointly considers the hidden state and the grounding signal to make the final decision. Let $h_t \in \mathbb{R}^d$ denote the final hidden state of the frozen base model at decoding step~$t$.
We combine it with the grounding signal to form the method input $u_t = \operatorname{LN}([h_t;\, \log(1+m_t)])$,
where $\operatorname{LN}(\cdot)$ denotes layer normalization. 
The trigger head then predicts whether to enter \textsc{copy} or remain in ordinary generation (\textsc{gen}):
\begin{equation}
   p_t^{\mathrm{trig}} = \operatorname{softmax}(W_{\mathrm{trig}}u_t + b_{\mathrm{trig}}). 
\end{equation}
At inference time, copying is activated only when the trigger selects \textsc{copy}. Otherwise, decoding remains in \textsc{gen} and falls back to standard one-token generation from the base model.

\subsection{Span Proposal}
\label{span}

Once copy mode is activated, the model must decide how far to follow the current source continuation. 
A fixed copy length is too rigid for this purpose. 
If it is too small, copying must be repeated over many rounds, increasing both overhead and creating more opportunities for transition errors. 
If it is too large, the proposal is more likely to cross the correct boundary, making verification harder and increasing the risk of over-copy.
Recognizing that a fixed copy length is suboptimal, we instead predict a \emph{tentative} copy length adaptively from the current state and leave the final decision to the verifier, which can still truncate the proposal if necessary.
To make this prediction stable under the highly skewed distribution of copy lengths, we predict length in log space. 
Specifically, using the same plugin input $u_t$ defined in \S\ref{trigger}, the length head $f_{\mathrm{len}}(\cdot)$ predicts a tentative copy length:
\begin{equation}
   \hat{\ell}_t
=
\left\lfloor
\exp \bigl(\operatorname{softplus}(f_{\mathrm{len}}(u_t))\bigr)
\right\rfloor .
\end{equation}
Here $\hat{\ell}_t$ is the proposed number of source tokens to copy from the current aligned continuation. 
Candidate source starts are drawn from the aligned continuation set $\mathcal{C}_t$, which together with the proposed length $\hat{\ell}_t$ defines the copy span to be verified.

\subsection{Proposal Verification}

\label{verification}

\begin{figure}
  \centering
  \includegraphics[width=1\textwidth]{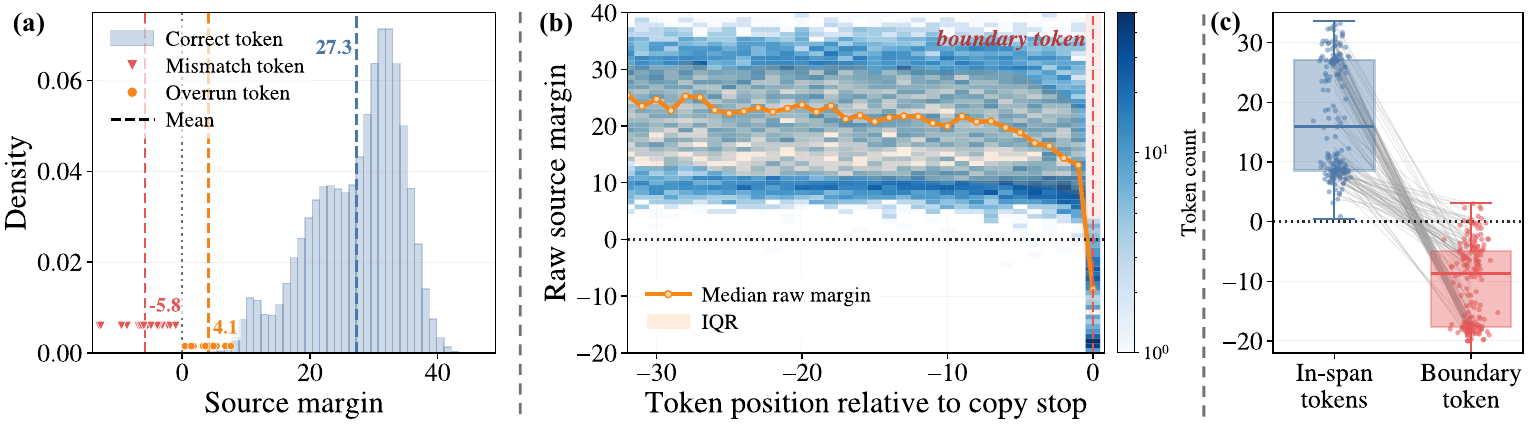}
\caption{
Source-margin behavior for Qwen3-8B.
(a) Correct copy tokens typically have large positive source margins, while mismatch and overrun cases lie close to the decision boundary.
(b) Around copy-stop positions, the raw source margin stays positive within the copied span and drops sharply at the boundary token.
(c) At the span level, in-span tokens tend to have much higher source margins than the boundary token.
}
\label{fig:logit_analysis}
\vspace{-3mm}
\end{figure}

Given a proposed source span, the verifier's role is to determine how much of it can be safely emitted. 
To inform this decision, we conduct an empirical analysis on output token logits during decoding, as shown in Figure~\ref{fig:logit_analysis}.
For a candidate source token $x_j$ and the frozen model's next-token logit vector $o$, we define the \textbf{\emph{source margin}} as
$\Delta=o[x_j]-\max_{v\in\mathcal{V},\,v\neq x_j}o[v]$,
i.e., the gap between the source-token logit and the strongest alternative-token logit.
A key observation is that most correctly copied tokens are preferred by a large margin, so these decisions are highly confident and far from the boundary. 
By contrast, copy errors concentrate near the decision boundary. 
Internal mismatches typically occur when the correct source token falls slightly \emph{below} the strongest competing token. 
Overrun errors, in contrast, often occur when the next source token remains only slightly \emph{above} the strongest alternative, even though generation should have left the copied span. 
This pattern suggests that the verifier should not replace the base model's preference altogether, but should apply a small adaptive correction precisely in these boundary cases.
We implement this through margin calibration, prefix acceptance, and a boundary-focused training objective.

\textbf{Margin calibration.}
Since errors concentrate near the decision boundary, we calibrate the raw source margin at each proposal token with a lightweight bias head.
For a candidate source start $s \in \mathcal{C}_t$, let $\pi_t^{(s)} = x_{s:s+\hat{\ell}_t-1}$ denote the proposed source span, and let $\pi_{t,k}^{(s)}$ denote its $k$-th token. 
We replay this proposal token by token with the frozen base model, obtaining logits
$o_{t,k}^{(s)} \in \mathbb{R}^{|\mathcal{V}|}$ at the $k$-th proposal position, and compute the raw source margin:
\begin{align}
    \Delta_{t,k}^{(s)} = o_{t,k}^{(s)}\!\bigl[\pi_{t,k}^{(s)}\bigr] - \textstyle \max_{v \in \mathcal{V},\, v \neq \pi_{t,k}^{(s)}} o_{t,k}^{(s)}[v].
\end{align}
A bias head $f_{\mathrm{bias}}(\cdot)$ computes a bias term
$b_{t,k}^{(s)} = \alpha \tanh\!\bigl(f_{\mathrm{bias}}(\operatorname{LN}([\bar{h}_{t,k}^{(s)};\Delta_{t,k}^{(s)}]))\bigr)$,
where $\bar{h}_{t,k}^{(s)}$ is the replay hidden state at the $k$-th proposal position and $\alpha$ sets the maximum correction magnitude.
We then calibrate the raw margin as
$\widetilde{\Delta}_{t,k}^{(s)} = \Delta_{t,k}^{(s)} + b_{t,k}^{(s)}$.
Intuitively, the bias head learns when to reinforce a source token and when to suppress it: for internal mismatch cases it can push an under-ranked but correct source token above its competitor, and for overrun cases it can push a marginally favored source token back below the alternative that should be generated instead.

\textbf{Prefix acceptance.}
Rather than accepting or rejecting a proposal wholesale, we emit only the longest prefix for which every token passes the margin threshold.
We define the accepted length as
\[
a_t^{(s)} = \max\Bigl\{L \in \{0,\dots,|\pi_t^{(s)}|\} : \widetilde{\Delta}_{t,k}^{(s)} > \gamma,\ \forall k=1,\dots,L\Bigr\},
\]
where $\gamma$ is the verifier threshold, and the accepted prefix is $A_t^{(s)}=\pi_t^{(s)}[1:a_t^{(s)}]$.
The decoder emits the accepted prefix of the best candidate in one step, truncating at the first unsafe token.
Candidates are ranked primarily by accepted length, with cumulative adjusted margin and cumulative log-probability as tie-breakers.
If no proposal is accepted, the decoder falls back to a one-step source-aware generation using the same bias head; details are provided in Appendix~\ref{appendix:verification}.

\textbf{Training objective.}
As observed in Figure \ref{fig:logit_analysis}, since the vast majority of correct copied tokens tend to have large positive source margins, we train the bias head with a hardness-weighted margin objective that focuses learning on ambiguous boundary cases. 
For each replayed position $i$, let $y_i \in \{+1,-1\}$ indicate whether it is an in-span copy token or a boundary-crossing token, and let $\Delta_i^{\mathrm{raw}}$ and $\Delta_i^{\mathrm{adj}}$ denote its raw and adjusted margins. We define a hardness weight as:
\begin{align}
    w_i^{\mathrm{hard}} = \sigma\!\left(-\textstyle \frac{y_i \Delta_i^{\mathrm{raw}}}{\tau_w}\right),
\end{align}
where $\sigma$ denotes the logistic sigmoid and $\tau_w$ is a temperature.
The main verifier objective is
\begin{align}
    \mathcal{L}_{\mathrm{margin}} =
    \frac{\sum_{i:y_i=+1} w_i^{\mathrm{hard}} [m_+ - \Delta_i^{\mathrm{adj}}]_+}
    {\sum_{i:y_i=+1} w_i^{\mathrm{hard}} + \varepsilon}
    + \lambda_-
    \frac{\sum_{i:y_i=-1} w_i^{\mathrm{hard}} [\Delta_i^{\mathrm{adj}} + m_-]_+}
    {\sum_{i:y_i=-1} w_i^{\mathrm{hard}} + \varepsilon},
\end{align}
where $m_+$ and $m_-$ are positive and negative safety margins, $[z]_+ = \max(z,0)$, $\lambda_-$ weights the negative term, and $\varepsilon$ is a small constant for stability. 
This encourages in-span copy tokens to remain above the decision boundary and boundary-crossing tokens to remain below it.

% The final objective is

% $$\mathcal{L} = \mathcal{L}_{\mathrm{trig}} + \lambda_{\mathrm{len}}\mathcal{L}_{\mathrm{len}} + \lambda_{\mathrm{bias}}\bigl(\mathcal{L}_{\mathrm{margin}} + \mathcal{L}_{\mathrm{amp}}\bigr)$$

Overall, the verifier turns copy acceptance into a calibrated decision process: high-confidence source tokens pass through unchanged, difficult in-span tokens can be rescued by a positive correction, and marginal boundary-crossing tokens can be suppressed before they trigger over-copy.
The full decoding procedure is summarized in Appendix~\ref{app:algorithm}.

\section{Experiments}
\label{sec:experiment}

\subsection{Experimental Setup}

\textbf{Benchmark Details.}
We evaluate exact span reproduction on three real-world domains: \textit{Contract}, \textit{Literature}, and \textit{Code}.
Across all three domains, the target spans are highly sensitive to boundaries, formatting, and character-level corruption, so even small deviations can make the output unusable.

\textit{Contract} studies \textit{structured legal clause extraction}, where the model must identify the requested clause(s) from contracts curated from SEC EDGAR filings and place them into a JSON-style structured output format.
\textit{Literature} asks for \textit{scientific evidence extraction}, requiring the model to return verbatim supporting spans from full-text articles collected from PubMed Central, including experimental settings, subject or sample descriptions, and statistical analysis statements.
\textit{Code} focuses on \textit{code explanation}, where the model must briefly explain a target snippet and then reproduce it exactly, using contexts drawn from widely used open-source Python repositories.
To further evaluate exact-copy reliability under stress, we additionally include \textit{Random Sequence}, where the model must precisely copy synthetic OpenSSH-style key sequences from the input, reflecting cases where encoded configuration values or credential-like identifiers must be copied exactly.
Full benchmark construction details, data sources, and dataset statistics are provided in Appendix~\ref{bench_details}.

\textbf{Evaluation Metrics.}
We evaluate both correctness and efficiency. For correctness, we use Exact Match (EM), which requires the generated output to match the target span exactly. 
For efficiency, we report decoding throughput in tokens per second (Tokens/s), and summarize relative speedup.

\textbf{Models.}
We evaluate on three representative LLMs from two model families and scales: Qwen3-4B, Qwen3-8B \citep{yang2025qwen3}, and Ministral-8B-Instruct (Ministral-8B hereafter) \citep{mistral2024ministral8b}. 
This setup tests whether exact-copy behavior and the effect of our method are consistent across model families and sizes.

\textbf{Baselines.}
For each base model, we report four configurations following the structure of Table~\ref{tab:main_results}:
(1) \textbf{Base Model}, i.e., the original frozen LLM with standard decoding;
(2) \textbf{\method}, which applies our decoding-time plugin to the frozen base model;
(3) \textbf{Base Model$_{\text{SFT}}$}, which applies supervised finetuning using LoRA \citep{hu2022lora};
and (4) \textbf{\method$_{\text{SFT}}$}, which combines our plugin with the SFT-adapted model.
This setup allows us to separate the effects of decoding-time control from those of parameter-efficient finetuning, and to test whether the two are complementary.

To examine both the accuracy and efficiency of \method{}, we further compare it against three families of baselines that exploit the source text during decoding, as reported in Table~\ref{tab:baseline_main}:
(1) \textbf{Lossless Source-Reuse Acceleration}, including LLMA~\citep{yang2023inference} and CopySpec~\citep{dumitru-etal-2025-copyspec}, which speeds up decoding while preserving the base model's output distribution;
(2) \textbf{Rule-Based Source Constraints}, including a PICARD-style~\citep{scholak-etal-2021-picard} decoder that masks generation to source-consistent continuations and a non-learned Longest-Copy rule that directly copies the longest matched source continuation;
and and (3) \textbf{Learned Copy Mechanisms}, represented by CopySpan~\citep{panthaplackel2021copy}, a span-level pointer-generator network trained on top of the frozen LM.
We also evaluate three frontier LLMs, GPT-4.1-nano~\citep{openai2025gpt41}, Gemini-2.5-Flash-Lite~\citep{comanici2025gemini}, and DeepSeek-V4-Pro~\citep{deepseek2026v4}, under the same protocol (Table~\ref{tab:api_models}).

\textbf{Implementation Details.}
For SFT baselines, we use LoRA with rank 16.
For \method, we freeze the backbone LM and train only the lightweight plugin (\method$_{\text{SFT}}$ applies the plugin on top of the SFT backbone). 
For all trainable components, we use dropout 0.05 and gradient accumulation 16. 
Optimization uses AdamW \citep{loshchilov2018decoupled} with a cosine learning-rate schedule. 
All experiments are run in BF16 precision on one NVIDIA RTX PRO 6000 GPU. 
To ensure reproducibility and make decoding behavior comparable across methods, we use deterministic decoding with temperature set to 0 in all experiments; for \method, proposal verification uses the positive-margin threshold $\gamma=0$.
% Additional training and inference hyperparameters are provided in Appendix~\ref{app:impl_details}.
Detailed hyperparameters and prompt templates are provided in Appendices~\ref{app:impl_details} and~\ref{app:prompts}, respectively.

\subsection{Main Results}

\begin{table*}[t]
\caption{Overall performance comparison. Average column shows the mean results across all four datasets.
For all metrics, higher scores indicate better performance. 
Better results within each corresponding comparison pair are highlighted in \textbf{bold} (two-tailed paired t-test, $p<0.05$).}
\centering
\renewcommand{\arraystretch}{1.2}
\begin{adjustbox}{max width=\linewidth}
\begin{tabular}{l *{5}{cc}} 
\toprule[1.5pt]
\multirow{2}{*}{\textbf{Method}} & \multicolumn{2}{c}{\textbf{Contract}} & \multicolumn{2}{c}{\textbf{Literature}} & \multicolumn{2}{c}{\textbf{Code}}  & \multicolumn{2}{c}{\textbf{Random Sequence}} & \multicolumn{2}{c}{\textbf{Average}} \\
\cmidrule(lr){2-3} \cmidrule(lr){4-5} \cmidrule(lr){6-7} \cmidrule(lr){8-9} \cmidrule(lr){10-11}
& \textbf{EM} & \textbf{Tokens/s} & \textbf{EM} & \textbf{Tokens/s} & \textbf{EM} & \textbf{Tokens/s} & \textbf{EM} & \textbf{Tokens/s} & \textbf{EM} & \textbf{Tokens/s} \\
\midrule[1.5pt]

% --- Model 1 ---
\rowcolor{gray!15} \multicolumn{11}{c}{\textit{\textbf{Qwen3-4B}}} \\ 
Base Model & 5.67 & 66.54 & 68.28 & 60.09 & 54.15 & 54.75 & 53.60 & 62.28 & 45.43 & 60.92 \\
\rowcolor{green!3} \textbf{\method} & \textbf{13.48} & \textbf{238.02} & \textbf{84.83} & \textbf{729.42} & \textbf{55.02} & \textbf{132.28} & \textbf{74.80} & \textbf{2939.15} & \textbf{57.03} & \textbf{1009.72} \\
\hline
Base Model$_{\text{SFT}}$ & 90.07 & 43.62 & 84.83 & 44.35 & 97.38 & 43.88 & 56.00 & 44.77 & 82.07 & 44.16 \\
\rowcolor{green!10} \textbf{\method$_{\text{SFT}}$} & \textbf{90.78} & \textbf{160.12} & \textbf{85.52} & \textbf{149.24} & \textbf{98.69} & \textbf{424.87} & \textbf{90.40} & \textbf{2143.36} & \textbf{91.35} & \textbf{719.40} \\
\midrule

% --- Model 2 ---
\rowcolor{gray!15} \multicolumn{11}{c}{\textit{\textbf{Qwen3-8B}}} \\ 
Base Model & 16.31 & 53.00 & 77.24 & 56.30 & 80.35 & 56.39 & 67.20 & 56.18 & 60.28 & 55.47 \\
\rowcolor{green!3} \textbf{\method} & \textbf{34.75} & \textbf{219.54} & \textbf{86.21} & \textbf{592.85} & \textbf{96.51} & \textbf{72.79} & \textbf{90.40} & \textbf{2483.23} & \textbf{76.97} & \textbf{842.10} \\
\hline
Base Model$_{\text{SFT}}$ & \textbf{96.45} & 43.21 & \textbf{94.48} & 43.92 & 97.82 & 42.52 & 90.00 & 43.62 & 94.69 & 43.32 \\
\rowcolor{green!10} \textbf{\method$_{\text{SFT}}$} & \textbf{96.45} & \textbf{159.33} & \textbf{94.48} & \textbf{234.16} & \textbf{98.25} & \textbf{326.87} & \textbf{96.80} & \textbf{1333.66} & \textbf{96.50} & \textbf{513.51} \\
\midrule

% --- Model 3 ---
\rowcolor{gray!15} \multicolumn{11}{c}{\textit{\textbf{Ministral-8B}}} \\ 
Base Model & 19.15 & 60.01 & 41.38 & 62.05 & 96.07 & 63.44 & 42.80 & 60.93 & 49.85 & 61.61 \\
\rowcolor{green!3} \textbf{\method} & \textbf{22.70} & \textbf{204.22} & \textbf{88.28} & \textbf{521.48} & \textbf{96.51} & \textbf{69.11} & \textbf{100.00} & \textbf{2609.59} & \textbf{76.87} & \textbf{851.10} \\
\hline
Base Model$_{\text{SFT}}$ & \textbf{98.58} & 49.42 & \textbf{92.41} & 49.43 & 99.56 & 49.48 & 98.00 & 49.97 & 97.14 & 49.58 \\
\rowcolor{green!10} \textbf{\method$_{\text{SFT}}$} & \textbf{98.58} & \textbf{178.88} & \textbf{92.41} & \textbf{529.07} & \textbf{100.00} & \textbf{400.70} & \textbf{100.00} & \textbf{1939.16} & \textbf{97.75} & \textbf{761.95} \\

\bottomrule[1.5pt]
\end{tabular}
\end{adjustbox}
\label{tab:main_results}
\vspace{-4mm}
\end{table*}

\begin{table*}[t]
\caption{Comparison with source-reuse and constrained-decoding baselines. Results are averaged over the three backbones (Qwen3-4B, Qwen3-8B, Ministral-8B); per-backbone results are in Table~\ref{tab:baseline_full}. Throughput of Longest-Copy is omitted (--) as it is not meaningful without model verification. Best results in each column are highlighted in \textbf{bold} (two-tailed paired t-test, $p<0.05$).}
\centering
\renewcommand{\arraystretch}{1.2}
\begin{adjustbox}{max width=\linewidth}
\begin{tabular}{l *{5}{cc}}
\toprule[1.5pt]
\multirow{2}{*}{\textbf{Method}} & \multicolumn{2}{c}{\textbf{Contract}} & \multicolumn{2}{c}{\textbf{Literature}} & \multicolumn{2}{c}{\textbf{Code}}  & \multicolumn{2}{c}{\textbf{Random Sequence}} & \multicolumn{2}{c}{\textbf{Average}} \\
\cmidrule(lr){2-3} \cmidrule(lr){4-5} \cmidrule(lr){6-7} \cmidrule(lr){8-9} \cmidrule(lr){10-11}
& \textbf{EM} & \textbf{Tokens/s} & \textbf{EM} & \textbf{Tokens/s} & \textbf{EM} & \textbf{Tokens/s} & \textbf{EM} & \textbf{Tokens/s} & \textbf{EM} & \textbf{Tokens/s} \\
\midrule[1.5pt]
Base Model & 13.71 & 59.85 & 62.30 & 59.48 & 76.86 & 58.19 & 54.53 & 59.80 & 51.85 & 59.33 \\
\rowcolor{gray!15} \multicolumn{11}{c}{\textit{\textbf{Lossless Source-Reuse Acceleration}}} \\
LLMA     & 13.71 & \textbf{238.60} & 62.30 & 493.40 & 76.86 & 109.70 & 54.53 & 288.27 & 51.85 & 282.49 \\
CopySpec & 13.71 & 183.80 & 62.30 & 301.23 & 76.86 & 111.77 & 54.53 & 171.77 & 51.85 & 192.14 \\
\rowcolor{gray!15} \multicolumn{11}{c}{\textit{\textbf{Rule-Based Copy Constraint}}} \\
PICARD-style & 0.00 & 54.27 & 62.53 & 54.73 & 89.37 & 52.73 & 54.53 & 180.17 & 51.61 & 85.48 \\
Longest-Copy & 0.00 & -- & 12.41 & -- & \textbf{92.87} & -- & 0.00 & -- & 26.32 & -- \\
\rowcolor{gray!15} \multicolumn{11}{c}{\textit{\textbf{Learned Copy Mechanism}}} \\
CopySpan & 3.31 & 46.78 & 19.54 & 419.88 & 85.01 & \textbf{336.48} & 20.40 & 1422.80 & 32.06 & 556.48 \\
\midrule
\rowcolor{green!8} \textbf{\method\ (Ours)} & \textbf{23.64} & 220.59 & \textbf{86.44} & \textbf{614.58} & 82.68 & 91.39 & \textbf{88.40} & \textbf{2677.32} & \textbf{70.29} & \textbf{900.97} \\
\bottomrule[1.5pt]
\end{tabular}
\end{adjustbox}
\label{tab:baseline_main}
\vspace{-4mm}
\end{table*}

\begin{wrapfigure}{r}{0.5\textwidth}
    \vspace{-0em}
    \centering
    \includegraphics[width=\linewidth]{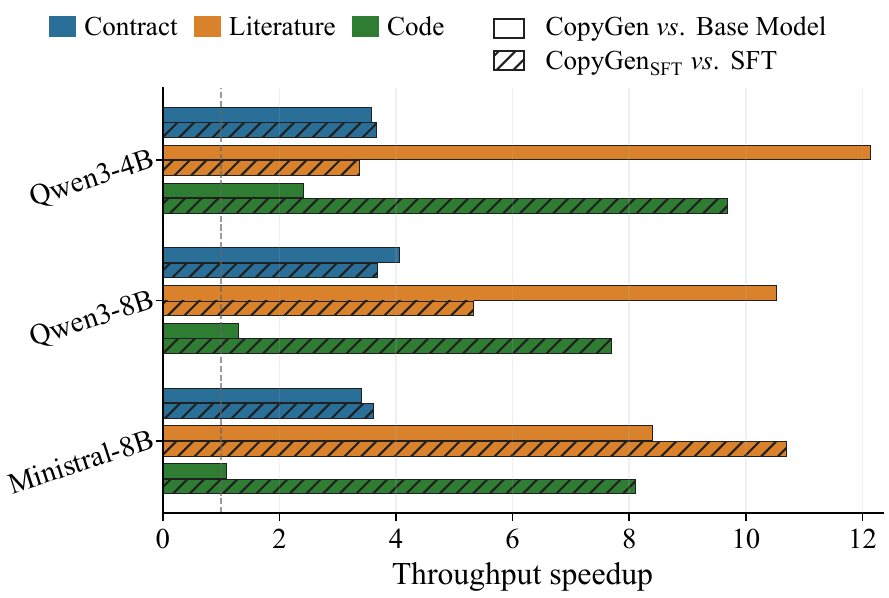}
    \caption{Throughput speedup on real-world tasks.}
    \label{fig:speedup}
    \vspace{-1.0em}
\end{wrapfigure}

Table~\ref{tab:main_results} reports exact-match accuracy and decoding throughput across four benchmark domains and three backbone LLMs, and Figure~\ref{fig:speedup} visualizes throughput gains on the three real-world tasks.
\textit{(1) \method\ substantially improves exact-copy fidelity and decoding efficiency for frozen LLMs.}
Averaged over the three frozen backbones, \method\ raises EM from $51.9$ to $70.3$ while increasing throughput from $59.3$ to $901.0$ tokens/s, corresponding to roughly a $15.2\times$ speedup. 
The gains are especially strong on \textit{Literature} and \textit{Random Sequence}, where long-span exact reproduction is most brittle under standard decoding. 
For example, the average EM improves from $60.3$ to $77.0$ on Qwen3-8B and from $49.9$ to $76.9$ on Ministral-8B, showing that \method\ strengthens exact copying without updating base model parameters.
\textit{(2) \method\ remains highly effective when combined with supervised finetuning.}
With SFT, \method\ further improves the average EM from $91.3$ to $95.2$, while increasing throughput from $45.7$ to $665.0$ tokens/s, a $14.6\times$ speedup. 
These results show that decoding-time copy control and parameter-efficient adaptation are complementary: finetuning improves the model's overall task performance, while \method\ further strengthens exact reproduction and substantially accelerates generation.
As shown in Figure~\ref{fig:speedup}, these speedups are consistent across real-world domains, reaching over $12\times$ in the base model setting and over $10\times$ in the SFT setting.
\textit{(3) The largest gains appear precisely on tasks where exact boundary and character-level fidelity matter most.}
On \textit{Contract}, \textit{Literature}, and \textit{Random Sequence}, even small local mistakes in punctuation, clause boundaries, or long low-redundancy strings invalidate the output, and \method\ yields the largest EM improvements. 
On \textit{Code}, the EM gains are more moderate but still consistent, while throughput improves substantially across all models. 
This pattern indicates that \method\ is effective at stabilizing long source-grounded spans and turning many fragile token-level decisions into fewer verified span emissions.
\textit{(4) \method\ is the only method that improves both accuracy and efficiency over source-reuse and constrained-decoding baselines.}
As shown in Table~\ref{tab:baseline_main}, lossless acceleration methods such as LLMA and CopySpec leave EM at the base-model level, whereas \method{} improves average EM by 18.44 points while running $3.19\times$ and $4.69\times$ faster than them.
Rule-based constraints and CopySpan all fall below the base model on average, since they cannot reliably determine where the requested span should start and stop in the source document, and the PICARD-style decoder even fails entirely on \textit{Contract}, where copied clauses must be embedded in JSON strings.
\textit{(5) Exact copying remains challenging even for frontier LLMs.}
Even the strongest frontier LLM, DeepSeek-V4-Pro, reaches only 56.03 EM on \textit{Contract} (Table~\ref{tab:api_models}). With \method{}, Qwen3-8B and Ministral-8B already surpass GPT-4.1-nano and Gemini-2.5-Flash-Lite without finetuning, and all three backbones outperform every frontier LLM when further combined with SFT.

\section{Analysis and Discussion}

\subsection{Ablation Study}

We also conduct an ablation study to examine the contribution of each key component in \method, as shown in Tables~\ref{tab:ablation_avg} and~\ref{tab:ablation_full}.
For \textit{w/o Length Proposal}, we replace the predicted tentative span length with a fixed copy length $k \in \{8,16,32,64,128,256\}$ and average results across the sweep. 
The throughput drop is the most pronounced, confirming that no single fixed value can consistently balance boundary precision and long-span efficiency across tasks. Shorter fixed lengths tend to be conservative and avoid over-copy, but they require more copy rounds and limit throughput gains, while longer ones improve speed at the cost of crossing copy boundaries more frequently. 
\begin{wraptable}{r}{0.40\columnwidth}
% \vspace{-2em}
\caption{Average ablation results of \method\ across the four benchmark domains. Best results are shown in \textbf{bold}.}
\centering
\small
\renewcommand{\arraystretch}{1.08}
\setlength{\tabcolsep}{3.5pt}
\begin{adjustbox}{max width=\linewidth}
\begin{tabular}{l l cc}
\toprule
\textbf{Model} & \textbf{Variant} & \textbf{EM ($\uparrow$)} & \textbf{Tokens/s ($\uparrow$)} \\
\midrule

\multirow{4}{*}{Qwen3-4B}
& \textbf{Full} & \textbf{57.03} & \textbf{1009.72} \\
& w/o Length Proposal & 56.69 & 452.49 \\
& w/o Adaptive Weighting & 46.78 & 690.55 \\
& w/o Margin Calibration & 44.94 & 798.14 \\
\midrule

\multirow{4}{*}{Qwen3-8B}
& \textbf{Full} & \textbf{76.97} & \textbf{842.10} \\
& w/o Length Proposal & 74.32 & 354.94 \\
& w/o Adaptive Weighting & 63.67 & 656.22 \\
& w/o Margin Calibration & 60.47 & 670.12 \\
\midrule

\multirow{4}{*}{Ministral-8B}
& \textbf{Full} & \textbf{76.87} & \textbf{851.10} \\
& w/o Length Proposal & 74.95 & 298.94 \\
& w/o Adaptive Weighting & 60.89 & 292.49 \\
& w/o Margin Calibration & 50.03 & 243.38 \\

\bottomrule
\end{tabular}
\end{adjustbox}
\label{tab:ablation_avg}
\vspace{-1.0em}
\end{wraptable}
When all replayed positions are assigned equal training weight (\textit{w/o Adaptive Weighting}), both exact match and throughput fall clearly, as learning becomes dominated by easy in-span copy tokens rather than the ambiguous near-boundary cases where copy failures actually occur.
The steepest degradation comes from removing margin calibration (\textit{w/o Margin Calibration}), which effectively reduces \method\ to copying only when the base model already favors the source token. 
In this case, the verifier can neither rescue under-ranked in-span tokens nor suppress marginal continuations near the true boundary, leaving the most error-prone decoding decisions unguarded and entirely to the frozen base model. 
Overall, length proposal drives efficiency, adaptive weighting focuses learning where it matters, and margin calibration is what ultimately turns that supervision into reliable exact-copy control.

\subsection{Copy Dynamics and Calibration}

\begin{figure}
    \centering
    \includegraphics[width=\linewidth]{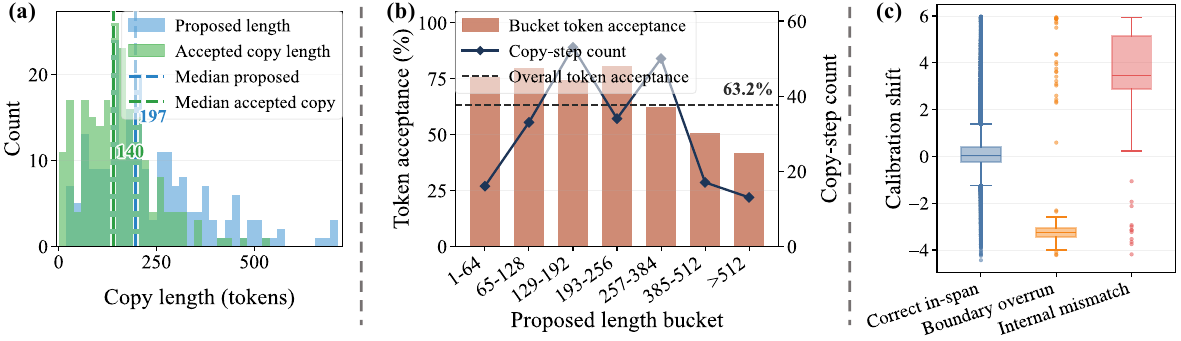}
\caption{Copy proposal dynamics and calibration behavior. (a) Proposed copy spans are often longer than the spans finally accepted by the verifier. (b) Token acceptance remains high for medium-length proposals but drops for very long proposals, indicating conservative truncation near uncertain boundaries. (c) Calibration bias suppresses boundary overruns while boosting internally fragile copy positions, helping preserve exact in-span reproduction. Results shown for Qwen3-8B.}
    \label{fig:copy_dynamics}
    \vspace{-3mm}
\end{figure}

We analyze the copy behavior of \method{} to understand whether it simply copies longer spans or learns to control when copying should stop. 
Using the literature domain as an example, Figure~\ref{fig:copy_dynamics}(a) shows that \method{} follows a propose-then-verify pattern: the median proposed length is 197 tokens, whereas the median accepted length is 140 tokens. 
This gap shows that \method{} often proposes broad source continuations and lets the verifier trim them before emission.
The results grouped by proposal length further reveal where this strategy is most useful. 
As shown in Figure~\ref{fig:copy_dynamics}(b), copy-step count rises from short to medium-length proposals and then drops for very long proposals, suggesting that most effective copy events occur at moderate span lengths. 
At the same time, token acceptance remains high for medium-length proposals but becomes more conservative for very long ones, indicating that verification mainly intervenes when boundary uncertainty increases.
The same proposal-and-acceptance trends are observed across other tasks and backbones, with full results provided in Appendix~\ref{app:copy_analysis}.
Finally, the calibration analysis in Figure~\ref{fig:copy_dynamics}(c) provides token-level evidence for this control mechanism. 
Confident in-span tokens receive little shift, preserving the base model's decisions. 
Boundary overrun tokens are pushed downward to prevent over-copying, while internally fragile positions receive positive bias to keep generation aligned with the source.

This analysis suggests that \method{} improves exact copying not by blindly copying more, but by combining broad span proposals with boundary-aware verification and local calibration.

\subsection{Integration with Inference Engines}
\label{sec:deployment}

\begin{wraptable}{r}{0.5\textwidth}
\vspace{-1em}
\centering
\caption{\method{} on Qwen3-8B under standard decoding and inference engines, averaged over four domains (full results in Table~\ref{tab:engine}). Relative EM changes are computed against the base model; speedup is the mean of per-domain speedups.}
\label{tab:deployment}
\vspace{2pt}
\small
\renewcommand{\arraystretch}{1.15}
\setlength{\tabcolsep}{3pt}
\begin{adjustbox}{max width=\linewidth}
\begin{tabular}{l cc cc c}
\toprule
\multirow{2}{*}{\textbf{Setting}} & \multicolumn{2}{c}{\textbf{EM ($\uparrow$)}} & \multicolumn{2}{c}{\textbf{Tokens/s ($\uparrow$)}} & \multirow{2}{*}{\textbf{Speedup}} \\
\cmidrule(lr){2-3} \cmidrule(lr){4-5}
& \textbf{Base} & \textbf{\method} & \textbf{Base} & \textbf{\method} & \\
\midrule
Standard & 60.28 & \textbf{76.97} (+27.7\%) & 55.47  & \textbf{842.10}  & $15.04\times$ \\
vLLM     & 60.69 & \textbf{77.00} (+26.9\%) & 374.57 & \textbf{2342.67} & $5.74\times$ \\
SGLang   & 60.67 & \textbf{77.06} (+27.0\%) & 366.47 & \textbf{2074.49} & $5.78\times$ \\
\bottomrule
\end{tabular}
\end{adjustbox}
\vspace{-1em}
\end{wraptable}

In real-world deployment, LLMs are typically served through optimized inference engines, and the practical value of \method{} ultimately depends on whether its gains carry over to such systems. We therefore integrate \method{} with Qwen3-8B into vLLM~\citep{vllm} and SGLang~\citep{zheng2024sglang}.
As shown in Table~\ref{tab:deployment}, \method{} improves EM by about 27\% relative in both engines, matching its gain under standard decoding, and raises throughput to over 2{,}000 tokens/s, corresponding to average speedups of $5.74\times$ and $5.78\times$. These speedups are obtained on top of engines that already apply batching and highly optimized execution, while the plugin heads add less than $2.7\%$ to wall-clock time, indicating that \method{} can be readily and seamlessly integrated into existing serving stacks.

\subsection{Training Efficiency Analysis}

\begin{figure}[t]
    \centering
    \begin{minipage}[t]{0.54\textwidth}
        \vspace{0pt}
        \centering
        \includegraphics[width=\linewidth]{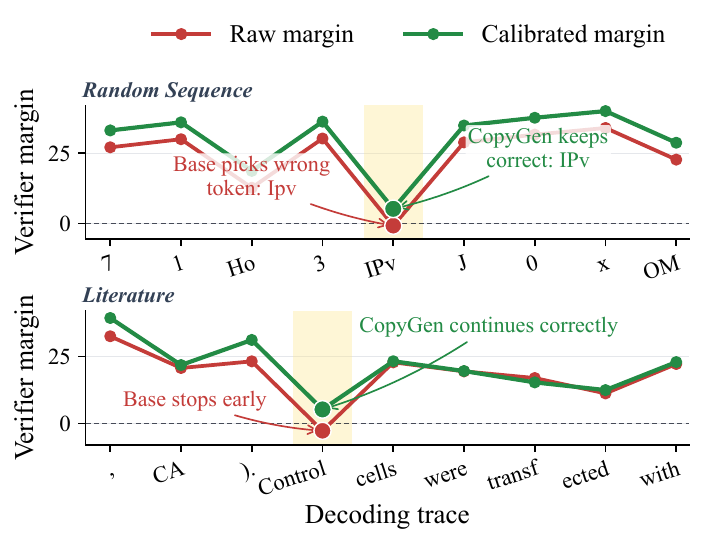}
        \caption{Calibrated decoding trace case study.}
        \label{fig:case_study}
    \end{minipage}
    \hfill
    \begin{minipage}[t]{0.43\textwidth}
        \vspace{0pt}
        \centering
        \captionof{table}{Average training resource comparison across four tasks. Relative changes are computed against SFT.}
        \label{tab:train_resource_avg}
        \small
        \setlength{\tabcolsep}{3.2pt}
        \begin{adjustbox}{max width=\linewidth}
        \begin{tabular}{lcc}
        \toprule
        \textbf{Method} & \textbf{Peak GiB ($\downarrow$)} & \textbf{Train s/item ($\downarrow$)} \\
        \midrule

        \rowcolor{gray!15} \multicolumn{3}{c}{\textit{\textbf{Qwen3-4B}}} \\
        SFT & 27.25 & 1.935 \\
        \rowcolor{green!3} \textbf{CopyGen} & \textbf{10.95} (-59.8\%) & \textbf{1.608} (-16.9\%) \\
        \rowcolor{green!10} \textbf{\method$_{\text{SFT}}$} & 11.00 (-59.6\%) & 1.724 (-10.9\%) \\
        \midrule

        \rowcolor{gray!15} \multicolumn{3}{c}{\textit{\textbf{Qwen3-8B}}} \\
        SFT & 35.55 & 1.597 \\
        \rowcolor{green!3} \textbf{CopyGen} & \textbf{19.01} (-46.5\%) & 1.600 (+0.2\%) \\
        \rowcolor{green!10} \textbf{\method$_{\text{SFT}}$} & 19.06 (-46.4\%) & \textbf{1.590} (-0.5\%) \\
        \midrule

        \rowcolor{gray!15} \multicolumn{3}{c}{\textit{\textbf{Ministral-8B}}} \\
        SFT & 32.51 & 1.639 \\
        \rowcolor{green!3} \textbf{CopyGen} & \textbf{18.65} (-42.6\%) & 1.649 (+0.6\%) \\
        \rowcolor{green!10} \textbf{\method$_{\text{SFT}}$} & 18.68 (-42.5\%) & \textbf{1.600} (-2.4\%) \\
        \bottomrule
        \end{tabular}
        \end{adjustbox}
    \end{minipage}
    \vspace{-2mm}
\end{figure}

Beyond improving decoding efficiency at inference time, \method{} is also lightweight to train. 
Table~\ref{tab:train_resource_avg} summarizes the training resources required by SFT and \method{} across the three backbone models, with full statistics in Table~\ref{tab:train_resource}. 
Since \method{} freezes the base LM and trains only small plugin heads, it substantially reduces peak GPU memory relative to LoRA SFT across all backbones, making copy-specific adaptation practical even under tight hardware budgets. 
Despite the lower memory usage, training time per example remains broadly comparable to SFT, with marginal improvements in several settings. 
Notably, these efficiency and accuracy gains come without touching the base model parameters, making \method{} a low-overhead, plug-and-play addition to existing LLMs.

\subsection{Case Study}
To intuitively understand how \method{} intervenes at the token level, we randomly select two concrete decoding failures and trace what calibration does to fix them.
In the first example, the model is copying a random key sequence and stumbles on a single character. 
The base model produces \texttt{...Gcw71Ho3\errorspan{Ipv}J0xOMk...}, swapping the capital \texttt{P} for a lowercase \texttt{p} mid-span. This is not a boundary error: the model is well inside an active copy region, and every surrounding token is decoded correctly. The top panel of Figure~\ref{fig:case_study} shows why it happens: the raw verifier margin for \texttt{IPv} dips just below zero at that position, and the base model tips toward the wrong token. Calibration pushes the margin back above the threshold, and \method{} emits the correct \texttt{...Gcw71Ho3\correctspan{IPv}J0xOMk...} without interruption.
The second example is a different kind of failure. 
Asked to extract the full transfection procedure from a scientific article, the base model stops prematurely after the first sentence, treating the period in ``...manufacturer instructions (Invitrogen, Carlsbad, CA).'' as a natural endpoint. 
The bottom panel shows the raw margin turning negative at \texttt{Control}, the token that should begin the next sentence. 
The base model reads this dip as a stopping cue, but \method{} recognizes it as a fragile transition rather than a true boundary and lifts the margin above zero, allowing decoding to continue into ``Control cells were transfected with empty vector.''

The two cases capture opposite failure modes, one where the model loses track mid-span, and one where it gives up too early.
Calibrated verification addresses both by learning to distinguish boundaries from low-confidence positions, rather than deferring to the base model's preference.

\section{Conclusion}

In this work, we study exact long-span reproduction as a distinct and practically important capability of large language models. 
Even when the target span is explicitly present in the input, standard decoding often fails to reproduce it faithfully, especially inside copied spans and near their boundaries. 
We introduce a benchmark for exact copying to make this problem concrete, and find current LLMs fall short in reproducing text exactly.
To address this problem, we propose \method{}, a copy-constrained plugin for frozen LLMs that decomposes exact copying into triggering, tentative span proposal, and token-level verification.
By calibrating source-continuation tokens during verification, experimental results show \method{} improves both exact-match accuracy and decoding efficiency across models and tasks, suggesting that reliable exact copying is not a trivial consequence of next-token prediction, but a decoding-time control problem that deserves explicit modeling. 
We hope this work encourages further research on reliable text generation with LLMs.

\bibliographystyle{abbrvnat}
\bibliography{references}

\clearpage

\appendix

\section{Limitations}

\label{app:limitations}
% Our study has several limitations. First, our benchmark focuses on exact span reproduction, so it mainly captures boundary fidelity and character-level correctness rather than broader forms of faithful generation such as paraphrasing or abstraction. Second, although we include realistic domains including contracts, scientific literature, and code, the current benchmark still covers only a limited range of exact-copy scenarios and does not consider settings such as noisy documents, OCR errors, or multilingual copying. Third, our method is a decoding-time plugin for frozen LLMs, which makes it lightweight but also means that its performance still depends on the underlying base model. If the base model fails to produce a reliable source-grounded continuation, the plugin has limited ability to recover. Finally, the efficiency gains of our method are task-dependent: when copy proposals are short or frequently rejected, the overhead of verification may reduce the practical speed advantage.

This study has several limitations.

First, our benchmark focuses on exact span reproduction, and therefore primarily captures boundary fidelity and character-level correctness rather than broader forms of faithful generation such as paraphrasing or abstraction. We believe that treating character-level exact copying as a standalone evaluation dimension is worthwhile in its own right. In many practical applications, semantic approximation is simply not enough, and this capability has long lacked dedicated attention in the literature, making our work a meaningful step toward filling that gap.

Second, although we include realistic domains such as contracts, scientific literature, and code, the current benchmark still covers only a limited range of exact-copy scenarios and does not consider settings such as noisy documents, OCR errors, or multilingual copying. We hope the three domains we selected serve as a representative starting point that can inform broader benchmark extensions in future work, rather than an attempt to exhaustively cover every possible scenario.

Third, our method is a decoding-time plugin for frozen LLMs, and its performance remains to some extent dependent on the underlying base model. This design trade-off is intentional. We aimed for CopyGen to integrate with existing models with minimal intrusiveness, and our experiments show that it consistently delivers meaningful gains even when the base model is already quite capable.

Finally, the efficiency gains of our method depend on both the task and the implementation. When copy proposals are short or frequently rejected, the overhead of verification may reduce the practical speed advantage. In addition, our current engine integration operates outside fused CUDA kernels, so on-the-fly suffix grounding introduces branching overhead in the decoding loop. Although this overhead is already reflected in our reported measurements, kernel-level batched suffix matching could reduce it further and is left for future work. That said, long-span, low-redundancy exact copying is a common and particularly challenging setting in practice, and it is also where \method{}'s efficiency gains are most pronounced. This suggests a natural alignment between where our method is designed to help and where it delivers the most value.

\section{Broader Impact}
\label{app:broader_impact}

Reliable exact span reproduction can have positive impacts in applications where faithful reuse of source text is important. For example, it can help code assistants preserve identifiers, configuration values, and code snippets exactly. It can also support legal and scientific workflows that require verbatim evidence extraction, while reducing errors introduced by paraphrasing or token-level regeneration.
Since \method{} can emit verified copied spans more efficiently, it may also reduce inference cost for copy-heavy generation tasks.

At the same time, improving exact copying can introduce risks if used without appropriate safeguards. A system with stronger copying ability may reproduce sensitive strings, private identifiers, copyrighted text, or other protected content when such content appears in the input context. It may also make users overconfident that copied outputs are always appropriate to reuse, even when the source material is confidential, legally restricted, or incorrectly selected. These risks are especially relevant in settings involving credentials, proprietary documents, legal text, or scientific claims.

Our work is intended for source-grounded copying where the target content is explicitly provided in the input and the user or application has permission to use it. We do not train models to extract memorized content from parameters, and the random-sequence benchmark uses only synthetic credential-like strings rather than real secrets. In practical deployments, exact-copy mechanisms should be combined with access control, data-permission checks, and safeguards against copying sensitive or restricted content.

\section{Algorithm Framework of \method}

\label{app:algorithm}

\begin{algorithm}[t]
\caption{Copy-Constrained Decoding of \method}
\label{alg:copygen}
\begin{algorithmic}[1]
\State \textbf{Input:} source $x_{1:M}$, instruction $r$, frozen LM $f_\theta$, plugin heads, thresholds $\tau,\gamma$
\State \textbf{Output:} generated output $y$
\State Build prompt $\mathcal{P}(x_{1:M},r)$ and initialize $y \leftarrow \emptyset$
\While{termination not reached}
    \State Compute hidden state $h_t$ and next-token logits $o_t$
    \If{$\arg\max o_t = \textsc{eos}$}
        \State \textbf{break}
    \EndIf
    \State Compute grounding signal $(m_t,\mathcal{C}_t)$ \Comment{longest grounded suffix and aligned continuations}
    \If{$m_t < \tau$}
        \State Emit $\arg\max o_t$ and continue
    \EndIf
    \State Form $u_t = \operatorname{LN}([h_t;\log(1+m_t)])$ and predict trigger $p_t^{\mathrm{trig}}$
    \If{$\arg\max p_t^{\mathrm{trig}}=\textsc{gen}$}
        \State Emit $\arg\max o_t$ and continue
    \EndIf
    \State Predict tentative copy length $\hat{\ell}_t$ and construct candidate starts $\mathcal{S}_t \subseteq \mathcal{C}_t$
    \ForAll{$s \in \mathcal{S}_t$}
        \State Form proposal $\pi_t^{(s)} = x_{s:s+\hat{\ell}_t-1}$
        \State Replay $\pi_t^{(s)}$ and compute calibrated margins $\widetilde{\Delta}_{t,k}^{(s)}$
        \State Compute accepted length $a_t^{(s)} = \max\{L:\widetilde{\Delta}_{t,k}^{(s)}>\gamma,\forall k=1,\dots,L\}$
        \State Set accepted prefix $A_t^{(s)}=\pi_t^{(s)}[1:a_t^{(s)}]$
    \EndFor
    \State Select the best candidate $s^\star$
    \If{$a_t^{(s^\star)} > 0$}
        \State Emit $A_t^{(s^\star)}$ \Comment{longest verified prefix}
    \Else
        \State Emit one source-aware fallback token
    \EndIf
\EndWhile
\State \textbf{return} $y$
\end{algorithmic}
\end{algorithm}

Algorithm \ref{alg:copygen} shows the detailed procedure of how \method{} works.
At each step, the decoder first measures whether the current prefix is sufficiently grounded in the source.
If so, it may enter copy mode, propose a tentative source span, and verify it token by token.
Only the longest verified prefix is emitted.
If no proposal token is accepted, the decoder performs a source-aware fallback generation step by reranking candidate source-follow tokens before emitting one token.

\section{Further Analyses}

\subsection{Fixed Copy Length}
\begin{figure}[h]
    \centering
    \includegraphics[width=\textwidth]{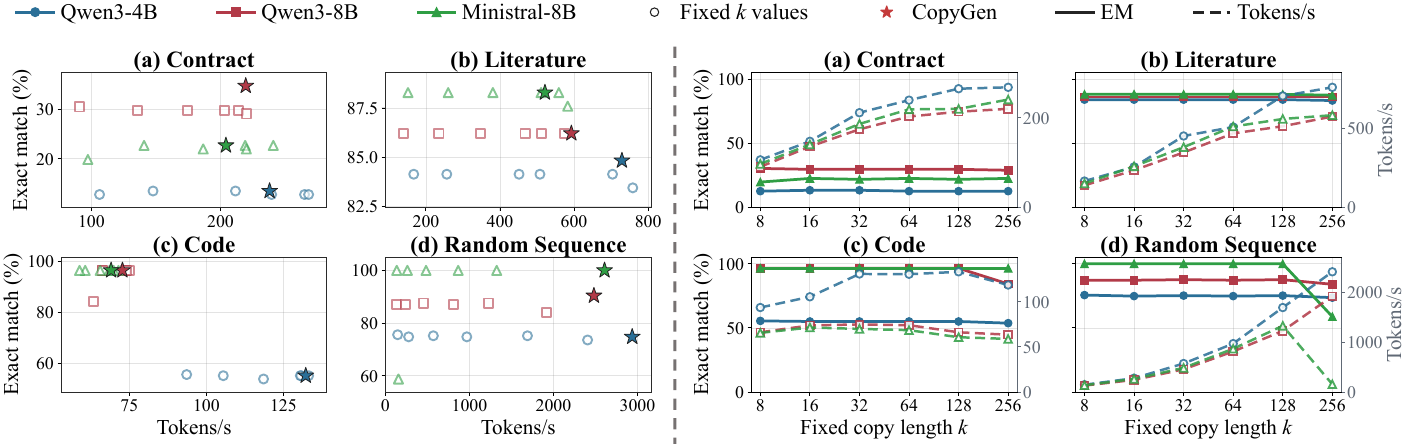}
        \caption{
    Fixed copy length analysis.
    Left: speed--accuracy tradeoff between fixed-length copying and \method{}'s adaptive span proposal, shown separately for each task.
    \method{} is close to the best fixed-length tradeoffs, preserving the accuracy of shorter fixed spans while approaching the throughput of longer fixed spans.
    Right: fixed-length sweep over $k$, where solid lines denote exact match and dashed lines denote throughput.
    Small fixed lengths preserve accuracy but limit throughput, while large fixed lengths improve throughput but can cross copy boundaries and hurt accuracy.
    }
        \label{fig:fixed-copy-length-sweep}
    \end{figure}

To isolate the effect of adaptive span proposal, we compare \method{} with variants that replace the predicted tentative span length by a fixed global copy length $k$.
We sweep $k \in \{8,16,32,64,128,256\}$ while keeping the remaining decoding and verification procedure unchanged.
Figure~\ref{fig:fixed-copy-length-sweep} shows two complementary views.
The left panel plots the speed--accuracy tradeoff for each task.
Across tasks, \method{} is consistently close to the best fixed-length tradeoffs: it preserves the accuracy of shorter fixed lengths while achieving throughput closer to longer fixed lengths.
The right panel shows the fixed-length sweep directly as a function of $k$.
The results show that no single fixed length is consistently best across tasks and models.
Smaller values of $k$ are conservative and help avoid over-copy, but they require more copy rounds and provide limited throughput gains.
Larger values of $k$ increase throughput, but they are more likely to cross the correct copy boundary and can reduce exact match.
\method{} avoids this global tuning problem by predicting a tentative length from the current decoding state and leaving the final boundary decision to verification.
This adaptive proposal lets \method{} achieve a stronger speed--accuracy tradeoff than choosing a single fixed copy length.

\subsection{Domain Generalization}
% \textbf{Domain Generalization Analysis.}
% OOD
We further evaluate whether \method\ generalizes across real-world domains using a leave-one-domain-out setup on Qwen3-8B and Ministral-8B. 
For each target domain, \method\ is trained on the other two real-world domains and evaluated directly on the held-out one. 
As shown in Table~\ref{tab:ood_generalization}, the held-out-domain variant performs better than the base model overall, while trailing the corresponding in-domain \method\ result.
This indicates that \method\ has nontrivial cross-domain generalization ability: the learned verifier transfers to unseen real-world domains and still delivers clear gains over the base model, although in-domain data remains helpful for best performance.

\begin{table*}[h]
\caption{Cross-domain generalization of \method\ under leave-one-domain-out training on the three real-world domains. \method\ (Held-out) trains on the two non-target real-world domains and evaluates on the held-out domain. Average is computed over Contract, Literature, and Code. Best results are highlighted in \textbf{bold}.}
\centering
\renewcommand{\arraystretch}{1.15}
\setlength{\tabcolsep}{4pt}
\begin{adjustbox}{max width=\linewidth}
\begin{tabular}{l l cc cc cc}
\toprule[1.5pt]
\multirow{2}{*}{\textbf{Test Domain}} & \multirow{2}{*}{\textbf{Train Domains}}
& \multicolumn{2}{c}{\textbf{Base Model}}
& \multicolumn{2}{c}{\textbf{\method}}
& \multicolumn{2}{c}{\textbf{\method\ (Held-out)}} \\
\cmidrule(lr){3-4} \cmidrule(lr){5-6} \cmidrule(lr){7-8}
& & \textbf{EM ($\uparrow$)} & \textbf{Tokens/s ($\uparrow$)}
& \textbf{EM ($\uparrow$)} & \textbf{Tokens/s ($\uparrow$)}
& \textbf{EM ($\uparrow$)} & \textbf{Tokens/s ($\uparrow$)} \\
\midrule[1.5pt]

\rowcolor{gray!15} \multicolumn{8}{c}{\textit{\textbf{Qwen3-8B}}} \\
Contract   & Literature + Code      & 16.31 & 53.00 & \textbf{34.75} & \textbf{219.54} & 16.31 & 162.05 \\
Literature & Contract + Code        & 77.24 & 56.30 & \textbf{86.21} & \textbf{592.85} & 85.52 & 391.49 \\
Code       & Contract + Literature  & 80.35 & 56.39 & \textbf{96.51} & \textbf{72.79} & 94.76 & 68.01 \\
\textbf{Average}    & --                     & 57.97 & 55.23 & \textbf{72.49} & \textbf{295.06} & 65.53 & 207.18 \\
\midrule

\rowcolor{gray!15} \multicolumn{8}{c}{\textit{\textbf{Ministral-8B}}} \\
Contract   & Literature + Code      & 19.15 & 60.01 & \textbf{22.70} & \textbf{204.22} & 21.28 & 179.81 \\
Literature & Contract + Code        & 41.38 & 62.05 & \textbf{88.28} & \textbf{521.48} & 68.97 & 393.91 \\
Code       & Contract + Literature  & 96.07 & 63.44 & \textbf{96.51} & 69.11 & 96.07 & \textbf{69.70} \\
\textbf{Average}    & --                     & 52.20 & 61.83 & \textbf{69.16} & \textbf{264.94} & 62.11 & 214.47 \\
\bottomrule[1.5pt]
\end{tabular}
\end{adjustbox}
\label{tab:ood_generalization}
\end{table*}

\subsection{Data Efficiency}
To examine whether \method\ remains effective under limited training data, we vary the amount of plugin training data on Qwen3-8B in the frozen-model setting, using 50\% and 10\% of the full training set. Table~\ref{tab:data_amount} shows that \method\ remains clearly better than the base model even under substantially reduced supervision. 
Using only 10\% of the training data still yields clear gains on all four domains, while 50\% is already sufficient to recover most of the full-data performance and even surpass it in some cases. 
These results show that \method\ remains effective with relatively limited training data, while additional supervision is most helpful on the more challenging settings.

\begin{table*}[h]
\caption{Effect of training data size on Qwen3-8B with \method. Best results in \textbf{bold}.}
\centering
\renewcommand{\arraystretch}{1.15}
\begin{adjustbox}{max width=\linewidth}
\begin{tabular}{l *{5}{cc}}
\toprule[1.5pt]
\multirow{2}{*}{\textbf{Method}} 
& \multicolumn{2}{c}{\textbf{Contract}}
& \multicolumn{2}{c}{\textbf{Literature}}
& \multicolumn{2}{c}{\textbf{Code}}
& \multicolumn{2}{c}{\textbf{Random Sequence}}
& \multicolumn{2}{c}{\textbf{Average}} \\
\cmidrule(lr){2-3} \cmidrule(lr){4-5} \cmidrule(lr){6-7} \cmidrule(lr){8-9} \cmidrule(lr){10-11}
& \textbf{EM ($\uparrow$)} & \textbf{Tokens/s ($\uparrow$)}
& \textbf{EM ($\uparrow$)} & \textbf{Tokens/s ($\uparrow$)}
& \textbf{EM ($\uparrow$)} & \textbf{Tokens/s ($\uparrow$)}
& \textbf{EM ($\uparrow$)} & \textbf{Tokens/s ($\uparrow$)}
& \textbf{EM ($\uparrow$)} & \textbf{Tokens/s ($\uparrow$)} \\
\midrule[1.5pt]
Base Model & 16.31 & 53.00 & 77.24 & 56.30 & 80.35 & 56.39 & 67.20 & 56.18 & 60.28 & 55.47 \\
\midrule
\method-100\% & \textbf{34.75} & \textbf{219.54} & \textbf{86.21} & \textbf{592.85} & \textbf{96.51} & 72.79 & \textbf{90.40} & 2483.23 & \textbf{76.97} & 842.10 \\
\method-50\%  & \textbf{34.75} & \textbf{219.54} & 85.52 & 480.51 & 96.07 & \textbf{73.36} & \textbf{90.40} & 2483.23 & 76.69 & 814.16 \\
\method-10\%  & 21.99 & 166.21 & 84.83 & 564.28 & 89.08 & 55.35 & 87.60 & \textbf{2661.27} & 70.88 & \textbf{861.78} \\
\bottomrule[1.5pt]
\end{tabular}
\end{adjustbox}
\label{tab:data_amount}
\end{table*}

\subsection{Learned Calibration versus Fixed Margin Rules}
\label{app:heuristic}

To isolate what the learned bias head contributes beyond a simple rule, we replace it with two prespecified heuristics applied after five grounded tokens. The first scales the raw margin by $2$, and the second adds a constant $1$, which is not sign-equivalent to the raw margin under $\gamma=0$. The trigger and length heads are kept unchanged, and all experiments use Qwen3-8B with temperature $0$. As shown in Table~\ref{tab:heuristic_calib}, both rules yield only marginal gains over the base model, whereas the learned bias head improves average EM on the real-world domains by nearly 15 points. This is because mid-span drifts and true copy boundaries overlap in raw-margin range, so a fixed rule shifts both in the same direction, while distinguishing them requires the hidden-state context used by the bias head.

\begin{table}[h]
\caption{Learned calibration versus fixed margin rules on Qwen3-8B. Avg is computed over the three real-world domains. Best results are highlighted in \textbf{bold}.}
\centering
\small
\renewcommand{\arraystretch}{1.12}
\begin{adjustbox}{max width=\linewidth}
\begin{tabular}{l ccc c c}
\toprule
\textbf{Calibration} & \textbf{Contract} & \textbf{Literature} & \textbf{Code} & \textbf{Avg} & \textbf{Random Sequence} \\
\midrule
Base Model & 16.31 & 77.24 & 80.35 & 57.97 & 67.20 \\
Raw margin $\times 2$ & 16.31 & 77.24 & 80.79 & 58.11 & 67.60 \\
Raw margin $+1$ & 17.02 & 77.93 & 83.41 & 59.45 & 71.20 \\
\rowcolor{green!5} \textbf{Learned bias head (\method)} & \textbf{35.46} & \textbf{86.21} & \textbf{96.94} & \textbf{72.87} & \textbf{90.40} \\
\bottomrule
\end{tabular}
\end{adjustbox}
\label{tab:heuristic_calib}
\end{table}

\subsection{Robustness to Prompts and Decoding}
\label{app:robustness}

We examine whether the gains of \method{} depend on the evaluation template or the decoding strategy. All experiments in this subsection use Qwen3-8B and reuse the trained checkpoint of each domain without retraining.

\textbf{Prompt and output-format robustness.}
We modify only the test-time prompt, using two paraphrases with different wording and sentence structure (PA, PB) and a tagged-output format (Tag) that requires the answer to be wrapped in \texttt{<COPYGEN\_OUTPUT>} tags. As shown in Table~\ref{tab:prompt_robust}, \method{} consistently outperforms the base model under all nine unseen variants, with an average gain of $+13.04$ EM. On \textit{Contract} PA, the gain even widens from $+18.44$ under the original template to $+22.70$, which would not be expected if the learned heads were overfitted to a particular template.

\begin{table*}[h]
\caption{Robustness to unseen instruction paraphrases (PA, PB) and output format (Tag) on Qwen3-8B. \textit{Orig.} denotes the original evaluation template in Table~\ref{tab:main_results}. Avg is computed over the nine unseen variants. Best results are highlighted in \textbf{bold}.}
\centering
\renewcommand{\arraystretch}{1.15}
\begin{adjustbox}{max width=\linewidth}
\begin{tabular}{l cccc cccc cccc c}
\toprule[1.5pt]
\multirow{2}{*}{\textbf{Method}} & \multicolumn{4}{c}{\textbf{Contract}} & \multicolumn{4}{c}{\textbf{Literature}} & \multicolumn{4}{c}{\textbf{Code}} & \multirow{2}{*}{\textbf{Avg}} \\
\cmidrule(lr){2-5} \cmidrule(lr){6-9} \cmidrule(lr){10-13}
& \textit{Orig.} & \textbf{PA} & \textbf{PB} & \textbf{Tag} & \textit{Orig.} & \textbf{PA} & \textbf{PB} & \textbf{Tag} & \textit{Orig.} & \textbf{PA} & \textbf{PB} & \textbf{Tag} & \\
\midrule[1.5pt]
Base Model & 16.31 & 14.89 & 16.31 & 16.31 & 77.24 & 77.93 & 55.86 & 73.79 & 80.35 & 71.18 & 89.52 & 89.08 & 56.10 \\
\rowcolor{green!5} \textbf{\method} & \textbf{34.75} & \textbf{37.59} & \textbf{34.04} & \textbf{33.33} & \textbf{86.21} & \textbf{86.21} & \textbf{75.17} & \textbf{78.62} & \textbf{96.51} & \textbf{84.72} & \textbf{97.38} & \textbf{95.20} & \textbf{69.14} \\
\midrule
$\Delta$EM & +18.44 & +22.70 & +17.73 & +17.02 & +8.97 & +8.28 & +19.31 & +4.83 & +16.16 & +13.54 & +7.86 & +6.11 & +13.04 \\
\bottomrule[1.5pt]
\end{tabular}
\end{adjustbox}
\label{tab:prompt_robust}
\end{table*}

\textbf{Non-greedy decoding.}
We evaluate at temperature $0.7$ with top-$p{=}0.9$ or min-$p{=}0.05$ on the three real-world domains using paired seeds 42/43/44. Verification remains a deterministic margin decision, and sampling applies to ordinary generation and rejected-proposal fallback. As shown in Table~\ref{tab:sampling}, \method{} exceeds the base model under every sampler and seed, and all values stay close to the greedy results in Table~\ref{tab:main_results}.

\begin{table}[h]
\caption{Exact match under non-greedy decoding on Qwen3-8B (temperature 0.7). SD denotes standard deviation over seeds.}
\centering
\small
\renewcommand{\arraystretch}{1.12}
\begin{adjustbox}{max width=\linewidth}
\begin{tabular}{l l c c c}
\toprule
\textbf{Domain} & \textbf{Sampler} & \textbf{Base (s42/s43/s44)} & \textbf{\method\ (s42/s43/s44)} & \textbf{Gain $\pm$ SD} \\
\midrule
\multirow{2}{*}{Contract}   & top-$p$ & 16.31/16.31/17.02 & 35.46/36.17/34.04 & +18.68$\pm$1.48 \\
                            & min-$p$ & 16.31/16.31/17.02 & 35.46/36.17/34.75 & +18.91$\pm$1.08 \\
\midrule
\multirow{2}{*}{Literature} & top-$p$ & 76.55/76.55/75.86 & 86.90/86.21/87.59 & +10.57$\pm$1.05 \\
                            & min-$p$ & 76.55/76.55/76.55 & 86.90/86.21/87.59 & +10.34$\pm$0.69 \\
\midrule
\multirow{2}{*}{Code}       & top-$p$ & 80.79/80.35/81.22 & 96.94/96.07/96.07 & +15.57$\pm$0.67 \\
                            & min-$p$ & 80.79/81.22/80.79 & 96.07/96.51/96.07 & +15.28$\pm$0.00 \\
\bottomrule
\end{tabular}
\end{adjustbox}
\label{tab:sampling}
\end{table}

\textbf{Distribution distortion at and after copy exit.}
We further check whether the calibration head distorts the sampling distribution around the copy boundary. We teacher-force each example to the end of its gold answer so that the base model and \method{} reach an identical state, and compare their next-token distributions under both samplers. As shown in Table~\ref{tab:exit_dist}, the distributions are nearly identical, with a mean total variation below $0.002$ and no top-1 flips. To test whether distortion persists after exit, we append an instruction requesting a short, varied free continuation, teacher-force the correct closing bridge, and sample three seeds per task. Table~\ref{tab:post_exit} shows that all diversity statistics are identical between the two models, since the plugin heads are never invoked after a correct exit.

\begin{table}[h]
\caption{Next-token distribution distortion at the copy exit boundary. TV denotes total variation distance between base and \method{} distributions (0 means identical).}
\centering
\small
\renewcommand{\arraystretch}{1.12}
\begin{adjustbox}{max width=\linewidth}
\begin{tabular}{l c c c c}
\toprule
\textbf{Probe} & \textbf{Comparable states} & \textbf{Top-1 flips} & \textbf{Mean TV} & \textbf{p95 TV} \\
\midrule
Gold-answer boundary (3 real domains) & 98.4\% & 0 (3 after masking EOS) & 0.000285 & 0 \\
First actual exit (Code) & 95.6\% & 0 & 0.001904 & 0 \\
\bottomrule
\end{tabular}
\end{adjustbox}
\label{tab:exit_dist}
\end{table}

\begin{table}[h]
\caption{Post-exit sampling diversity. All statistics are identical between the base model and \method{}; higher diversity values indicate more varied continuations across seeds.}
\centering
\small
\renewcommand{\arraystretch}{1.12}
\begin{adjustbox}{max width=\linewidth}
\begin{tabular}{l c c c c c}
\toprule
\textbf{Sampler} & \textbf{Token-identical pairs} & \textbf{Heads invoked} & \textbf{Unique rate} & \textbf{Pairwise diff.} & \textbf{Token-Jaccard dist.} \\
\midrule
top-$p$ & 100\% & 0 & 0.8750 & 0.8542 & 0.3177 \\
min-$p$ & 100\% & 0 & 0.9583 & 0.9479 & 0.3845 \\
\bottomrule
\end{tabular}
\end{adjustbox}
\label{tab:post_exit}
\end{table}

\subsection{Integration with Inference Engines}
\label{app:engine}

We integrate \method{} with Qwen3-8B into vLLM v0.25.1~\citep{vllm} and SGLang v0.5.15.post1~\citep{zheng2024sglang}. In vLLM, the engine executes the target model while the control logic of \method{} runs outside it. In SGLang, \method{} is registered as a custom \texttt{SpeculativeAlgorithm} inside the native speculative scheduler, so that no client-side re-prefill is needed. All measurements use batch size 8 on a single RTX PRO 6000 GPU, and Table~\ref{tab:engine} reports per-domain results. To bound the cost of the plugin heads, we run each example twice on an identical token trajectory, once normally and once with all three heads forced on every decoding step. The difference accounts for at most 1.746\% of wall-clock time in vLLM and 2.696\% in SGLang, since the speedup of \method{} comes from reducing the number of decoding steps rather than making each step faster.

\begin{table*}[h]
\caption{End-to-end evaluation of \method{} in vLLM and SGLang on Qwen3-8B. Average speedup is the mean of per-domain speedups. Better results are highlighted in \textbf{bold}.}
\centering
\renewcommand{\arraystretch}{1.05}
\begin{adjustbox}{max width=\linewidth}
\begin{tabular}{l ccc ccc}
\toprule[1.5pt]
\multirow{2}{*}{\textbf{Domain}} & \multicolumn{3}{c}{\textbf{EM ($\uparrow$)}} & \multicolumn{3}{c}{\textbf{Tokens/s ($\uparrow$)}} \\
\cmidrule(lr){2-4} \cmidrule(lr){5-7}
& \textbf{Base} & \textbf{\method} & \textbf{$\Delta$} & \textbf{Base} & \textbf{\method} & \textbf{Speedup} \\
\midrule

\rowcolor{gray!15} \multicolumn{7}{c}{\texttt{vLLM}} \\
Contract   & 16.31 & \textbf{34.75} & +18.44 & 381.94 & \textbf{1294.97} & $3.39\times$ \\
Literature & 77.24 & \textbf{85.52} & +8.28  & 172.24 & \textbf{838.85}  & $4.87\times$ \\
Code       & 80.79 & \textbf{96.94} & +16.15 & 448.13 & \textbf{518.45}  & $1.16\times$ \\
Random     & 68.40 & \textbf{90.80} & +22.40 & 495.95 & \textbf{6718.40} & $13.55\times$ \\
\hline
Avg (3 real domains) & 58.11 & \textbf{72.40} & +14.29 & 334.10 & \textbf{884.09}  & $3.14\times$ \\
Avg (all 4 domains) & 60.69 & \textbf{77.00} & +16.31 & 374.57 & \textbf{2342.67} & $5.74\times$ \\
\midrule

\rowcolor{gray!15} \multicolumn{7}{c}{\texttt{SGLang}} \\
Contract   & 16.31 & \textbf{35.46} & +19.15 & 388.36 & \textbf{672.91}  & $1.73\times$ \\
Literature & 77.24 & \textbf{85.52} & +8.28  & 317.03 & \textbf{1667.93} & $5.26\times$ \\
Code       & 79.91 & \textbf{96.07} & +16.16 & 392.61 & \textbf{476.75}  & $1.21\times$ \\
Random     & 69.20 & \textbf{91.20} & +22.00 & 367.86 & \textbf{5480.38} & $14.90\times$ \\
\hline
Avg (3 real domains) & 57.82 & \textbf{72.35} & +14.53 & 366.00 & \textbf{939.20}  & $2.74\times$ \\
Avg (all 4 domains) & 60.67 & \textbf{77.06} & +16.39 & 366.47 & \textbf{2074.49} & $5.78\times$ \\

\bottomrule[1.5pt]
\end{tabular}
\end{adjustbox}
\label{tab:engine}
\end{table*}

\section{Benchmark Details}
\label{bench_details}

We construct four benchmark domains to evaluate exact span reproduction under realistic conditions. 
Together, they cover settings where even small deviations in copied text can make the output unusable: legal clauses whose wording and numbering must remain intact, scientific evidence that must be quoted verbatim, code spans whose formatting and symbols are execution-critical, and low-redundancy strings that cannot be recovered from semantics alone.

\paragraph{Contracts.}
In the contract domain, we focus on structured \textit{legal clause extraction}.
We collected 2,662 exhibit files from SEC EDGAR filings\footnote{\url{https://www.sec.gov/edgar/search/}} and extracted text from 2,468 of them.
Given a local contract window, the model extracts requested clauses and returns them in a JSON-style object.
The output must preserve the original legal text exactly, including clause headings, numbering, and enumerated subclauses.
This reflects legal drafting and review workflows where downstream systems need exact clause text rather than paraphrased summaries.

\paragraph{Literature.}
In the literature domain, we focus on \textit{scientific evidence extraction}.
We curated 868 full-text articles from PubMed Central (PMC)\footnote{\url{https://pmc.ncbi.nlm.nih.gov/}}, covering 13,206 article sections.
Given a local article context, the model returns the exact evidence span for a requested scientific detail, such as an experimental protocol, sample description, assay procedure, or statistical analysis statement.
This reflects scientific search and review workflows where users need faithful quotations from the original paper, since small wording changes can distort methodological or empirical claims.

\paragraph{Code.}
In the code domain, we focus on \textit{code explanation}, using several well-known open-source Python web and API repositories, including \href{https://github.com/fastapi/fastapi}{\texttt{FastAPI}}, \href{https://github.com/fastapi/sqlmodel}{\texttt{SQLModel}}, and \href{https://github.com/encode/starlette}{\texttt{Starlette}}, whose APIs and test code is realistic and standard, allowing us to focus on faithful reproduction rather than deep program reasoning.
Together, these repositories contain over 1,500 Python files, more than 130K non-blank lines of Python code, and 766 test files.
Given a local code context, the model briefly explains a target snippet and then reproduces it exactly.
The task emphasizes faithful reproduction rather than deep program reasoning, since many target snippets contain fragile structures such as nested dictionaries, string literals, and brackets.

\paragraph{Random Sequence.}
We additionally include an \textit{exact long-string reproduction} task to isolate faithful copy when the target has little semantic redundancy. 
Specifically, we generate realistic OpenSSH-style private-key files and ask the model to reproduce the key sequence exactly.
All values are generated for this study and do not correspond to real credentials.
The target strings are long, low-redundancy sequences containing line breaks, mixed-case letters, digits, and symbols such as \texttt{/} and \texttt{+}, reflecting software and infrastructure workflows where encoded configuration values, serialized artifacts, or credential-like identifiers must be copied exactly.

\begin{table}[h]
  \caption{Dataset statistics for each domain and task.}
  \label{benchmark_statistics}
  \centering
  \begin{tabular}{llccc}
    \toprule
    Domain & Task & \# Train & \# Test & \# Total \\
    \midrule
    Contract & Legal Clause Extraction & 658 & 141 & 799 \\
    Literature & Scientific Evidence Extraction & 712 & 145 & 857 \\
    Code & Code Explanation & 821 & 229 & 1050 \\
    Random Sequence & Exact Long-String Reproduction & 2250 & 250 & 2500 \\
    \bottomrule
  \end{tabular}
\end{table}

Table~\ref{benchmark_statistics} summarizes the domains, tasks, and corresponding train/test sizes.

\section{Proposal Verification}
\label{appendix:verification}

For a candidate source start $s \in \mathcal{C}_t$, let
$\pi_t^{(s)} = x_{s:s+\hat{\ell}_t-1}$
denote the proposed source span. We replay this proposal token by token with the frozen base model. For the $k$-th proposal token, let $o_{t,k}^{(s)} \in \mathbb{R}^{|\mathcal{V}|}$
be the replay logit vector, and let $\pi_{t,k}^{(s)}$ denote the proposed source token at that position. We define the raw source margin as
\begin{equation}
   \Delta_{t,k}^{(s)}
=
o_{t,k}^{(s)}\!\bigl[\pi_{t,k}^{(s)}\bigr]
-
\max_{v \in \mathcal{V},\, v \neq \pi_{t,k}^{(s)}} o_{t,k}^{(s)}[v].
\end{equation}
This margin measures how strongly the frozen model prefers the proposed source token over the strongest alternative at the same decoding step.

We then apply a lightweight bias head to calibrate this raw margin. Specifically, a bias head $f_{\mathrm{bias}}(\cdot)$ computes
\[
b_{t,k}^{(s)}
=
\alpha \tanh\!\Bigl(
f_{\mathrm{bias}}\bigl(\operatorname{LN}([\bar{h}_{t,k}^{(s)};\Delta_{t,k}^{(s)}])\bigr)
\Bigr),
\]
where $\bar{h}_{t,k}^{(s)}$ is the replay hidden state at the $k$-th proposal position and $\alpha$ bounds the correction magnitude. The adjusted verifier margin is then
\[
\widetilde{\Delta}_{t,k}^{(s)}
=
\Delta_{t,k}^{(s)} + b_{t,k}^{(s)}.
\]
Intuitively, the bias head learns when the source token should be reinforced and when it should be suppressed. For internal mismatch cases, it can push an under-ranked but correct source token above its competitor. For overrun cases, it can push a marginally favored source continuation token back below the alternative that should be generated instead.

A proposal is accepted only up to its longest verified prefix. We first define the accepted length as
\[
a_t^{(s)}
=
\max \Bigl\{
L \in \{0,\dots,|\pi_t^{(s)}|\}
:
\widetilde{\Delta}_{t,k}^{(s)} > \gamma,\ \forall k=1,\dots,L
\Bigr\},
\]
where $\gamma$ is the verifier threshold, and the accepted prefix is $A_t^{(s)}=\pi_t^{(s)}[1:a_t^{(s)}]$.
Thus, the decoder emits only the longest safe prefix and truncates the proposal at the first unsafe token. Among all candidates, we rank proposals primarily by accepted length, and use cumulative adjusted margin and cumulative log-probability as tie-breakers.

If the best candidate $s^\star$ satisfies $a_t^{(s^\star)} > 0$, the accepted prefix $A_t^{(s^\star)}$ is emitted in one step.
Otherwise, the decoder performs a one-step fallback generation. This fallback is also calibrated by the same bias mechanism rather than reverting to plain greedy decoding. Let
\[
\mathcal{V}^{\mathrm{src}}_t
=
\{x_j : j \in \mathcal{C}_t\}
\]
denote the unique next-source tokens supported by the currently aligned continuation set. For each $v \in \mathcal{V}^{\mathrm{src}}_t$, we define a fallback raw margin
\[
\Delta_t^{\mathrm{fb}}(v)
=
o_t[v] - \max_{u \in \mathcal{V},\, u \neq v} o_t[u],
\]
where $o_t$ is the frozen model's next-token logit vector at the current decoding step. We reuse the same bias head to obtain
\[
b_t^{\mathrm{fb}}(v)
=
\alpha \tanh\!\Bigl(
f_{\mathrm{bias}}\bigl(\operatorname{LN}([h_t;\Delta_t^{\mathrm{fb}}(v)])\bigr)
\Bigr),
\]
and form calibrated fallback logits
\[
\bar{o}_t[v]
=
\begin{cases}
o_t[v] + b_t^{\mathrm{fb}}(v), & v \in \mathcal{V}^{\mathrm{src}}_t,\\
o_t[v], & \text{otherwise}.
\end{cases}
\]
This makes fallback generation more source-aware: when the correct next token should still follow the source, it can be strengthened; when source-following should end, candidate source-continuation tokens can be suppressed.

We train the bias head using token-level verifier supervision. For each replayed verifier position $i$, let $y_i \in \{+1,-1\}$ indicate whether the proposal token is a positive in-span copy token or a negative boundary-crossing token, and let $\Delta_i^{\mathrm{raw}}$ and $\Delta_i^{\mathrm{adj}}$ denote its raw and adjusted margins. Since most verifier decisions are easy and far from the decision boundary, we focus learning on the ambiguous cases by defining a hardness weight
\[
w_i^{\mathrm{hard}}
=
\sigma\!\left(
-\textstyle \frac{y_i \Delta_i^{\mathrm{raw}}}{\tau_w}
\right),
\]
where $\sigma(\cdot)$ denotes the logistic sigmoid and $\tau_w$ is a temperature parameter. Hard cases near the raw decision boundary receive larger weight, while easy high-confidence cases receive smaller weight.

The main verifier objective is
\[
    \mathcal{L}_{\mathrm{margin}} =
    \frac{\sum_{i:y_i=+1} w_i^{\mathrm{hard}} [m_+ - \Delta_i^{\mathrm{adj}}]_+}
    {\sum_{i:y_i=+1} w_i^{\mathrm{hard}} + \varepsilon}
    + \lambda_-
    \frac{\sum_{i:y_i=-1} w_i^{\mathrm{hard}} [\Delta_i^{\mathrm{adj}} + m_-]_+}
    {\sum_{i:y_i=-1} w_i^{\mathrm{hard}} + \varepsilon},
\]
where $m_+ > 0$ and $m_- > 0$ are safety margins for positive and negative verifier decisions, respectively, $[z]_+ = \max(z,0)$, $\lambda_-$ weights the negative term, and $\varepsilon$ is a small constant for numerical stability. This objective encourages positive copy tokens to remain safely above the decision boundary and negative boundary-crossing tokens to remain below it.

The final objective is
\[
\mathcal{L}
=
\mathcal{L}_{\mathrm{trig}}
+
\lambda_{\mathrm{len}}\mathcal{L}_{\mathrm{len}}
+
\lambda_{\mathrm{bias}} \mathcal{L}_{\mathrm{margin}},
\]
where $\mathcal{L}_{\mathrm{trig}}$ and $\mathcal{L}_{\mathrm{len}}$ denote the trigger and span-length objectives, and $\lambda_{\mathrm{len}}$ and $\lambda_{\mathrm{bias}}$ weight the length and margin terms, respectively.
Overall, the verifier turns copy acceptance into a calibrated decision process: high-confidence source tokens pass through unchanged, difficult in-span tokens can be rescued by a positive correction, and marginal boundary-crossing tokens can be suppressed before they trigger over-copy.

\section{Implementation Details}
\label{app:impl_details}

For the LoRA baseline, we insert LoRA adapters into the attention projection modules of the base model, namely \texttt{q\_proj}, \texttt{k\_proj}, \texttt{v\_proj}, and \texttt{o\_proj}. 
We use LoRA with rank $16$, scaling factor $32$, and dropout $0.05$.
Training runs for one epoch with AdamW, a learning rate of $2\times 10^{-4}$, gradient accumulation of $16$, and a warmup ratio of $0.1$. 
Following standard instruction-tuning practice, the loss is computed only on the target output, with prompt tokens masked out. 
At inference time, all baselines use greedy decoding.

For \method, we freeze the backbone LM and train only the lightweight trigger, length, and verifier-calibration heads. 
Each head is implemented as a small multilayer perceptron over decoder hidden states, with LayerNorm, GELU activations, and dropout. 
Training runs for one epoch with AdamW, a learning rate of $10^{-4}$, a cosine learning-rate schedule, and gradient accumulation of $16$. 
The training objective combines trigger prediction, span-length prediction, and verifier calibration, with $\lambda_{\mathrm{len}}=1$ and $\lambda_{\mathrm{bias}}=1$. 
Copy supervision is activated only when the current prefix has sufficient source grounding, using $\tau=5$, and the length objective is applied to the tentative span prediction.
For verifier calibration, we use $m_+=m_-=1$, $\lambda_-=1$, hardness temperature $\tau_w=1$, numerical constant $\varepsilon=10^{-8}$, and maximum bias magnitude $\alpha=8$.
At inference time, \method\ uses deterministic decoding with temperature set to $0$. 
Proposal verification uses a verifier threshold of $\gamma=0$, and when no proposal is accepted, decoding falls back to a source-aware one-step generation rule.

Throughput is measured with a single inference process on an otherwise idle GPU, so that the reported decoding speed is not affected by concurrent workloads.

Before computing exact match, we apply the same preprocessing procedure to both model outputs and reference answers to remove harmless formatting differences rather than relax the correctness criterion. 
For all tasks, this preprocessing strips leading and trailing whitespace and removes empty lines. 
For the code task, we additionally ignore whitespace-only formatting differences, such as uniform global indentation shifts and extra or missing blank lines. 
All remaining token, boundary, and content mismatches are counted as errors.

\section{Full Experimental Results}

\subsection{Full Baseline Comparisons}
\label{app:baseline_full}

Table~\ref{tab:baseline_full} reports per-backbone results for all baselines in Table~\ref{tab:baseline_main}. \method{} achieves the best average EM and throughput on every backbone, and consistently obtains the highest EM on \textit{Contract}, \textit{Literature}, and \textit{Random Sequence}.

\begin{table*}[h]
\caption{Full comparison with source-reuse and constrained-decoding baselines. Baselines are grouped into lossless source-reuse acceleration (\textit{Lossless}), rule-based source constraints (\textit{Rule-based}), and learned copy mechanisms (\textit{Learned}); \method{} is listed separately. Each cell reports EM and tokens/s. Throughput of Longest-Copy is omitted (--) as it is not meaningful without model verification. Best results within each backbone are highlighted in \textbf{bold}.}
\centering
\renewcommand{\arraystretch}{1.12}
\begin{adjustbox}{max width=\linewidth}
\begin{tabular}{l l *{5}{cc}}
\toprule[1.5pt]
\multirow{2}{*}{\textbf{Type}} & \multirow{2}{*}{\textbf{Method}} & \multicolumn{2}{c}{\textbf{Contract}} & \multicolumn{2}{c}{\textbf{Literature}} & \multicolumn{2}{c}{\textbf{Code}}  & \multicolumn{2}{c}{\textbf{Random Sequence}} & \multicolumn{2}{c}{\textbf{Average}} \\
\cmidrule(lr){3-4} \cmidrule(lr){5-6} \cmidrule(lr){7-8} \cmidrule(lr){9-10} \cmidrule(lr){11-12}
& & \textbf{EM} & \textbf{Tokens/s} & \textbf{EM} & \textbf{Tokens/s} & \textbf{EM} & \textbf{Tokens/s} & \textbf{EM} & \textbf{Tokens/s} & \textbf{EM} & \textbf{Tokens/s} \\
\midrule[1.5pt]
% --- Qwen3-4B ---
\rowcolor{gray!15} \multicolumn{12}{c}{\textit{\textbf{Qwen3-4B}}} \\
\multicolumn{2}{l}{Base Model} & 5.67 & 66.5 & 68.28 & 60.1 & 54.15 & 54.8 & 53.60 & 62.3 & 45.43 & 60.9 \\
\rowcolor{gray!6} & LLMA & 5.67 & \textbf{260.9} & 68.28 & 517.1 & 54.15 & 139.6 & 53.60 & 223.4 & 45.43 & 285.3 \\
\rowcolor{gray!6} \multirow{-2}{*}{\textit{Lossless}} & CopySpec & 5.67 & 187.2 & 68.28 & 313.8 & 54.15 & 136.7 & 53.60 & 116.3 & 45.43 & 188.5 \\
 & PICARD-style & 0.00 & 52.0 & 67.59 & 52.5 & 84.28 & 50.2 & 53.60 & 231.0 & 51.37 & 96.4 \\
\multirow{-2}{*}{\textit{Rule-based}} & Longest-Copy & 0.00 & -- & 12.41 & -- & \textbf{86.46} & -- & 0.00 & -- & 24.72 & -- \\
\rowcolor{gray!6} \textit{Learned} & CopySpan & 0.71 & 54.5 & 10.34 & \textbf{741.8} & 82.97 & \textbf{440.7} & 1.60 & 2522.2 & 23.91 & 939.8 \\
\rowcolor{green!8} \multicolumn{2}{l}{\textbf{\method\ (Ours)}} & \textbf{13.48} & 238.0 & \textbf{84.83} & 729.4 & 55.02 & 132.3 & \textbf{74.80} & \textbf{2939.2} & \textbf{57.03} & \textbf{1009.7} \\
\midrule
% --- Qwen3-8B ---
\rowcolor{gray!15} \multicolumn{12}{c}{\textit{\textbf{Qwen3-8B}}} \\
\multicolumn{2}{l}{Base Model} & 16.31 & 53.0 & 77.24 & 56.3 & 80.35 & 56.4 & 67.20 & 56.2 & 60.28 & 55.5 \\
\rowcolor{gray!6} & LLMA & 16.31 & \textbf{234.6} & 77.24 & 493.7 & 80.35 & 94.8 & 67.20 & 588.0 & 60.28 & 352.8 \\
\rowcolor{gray!6} \multirow{-2}{*}{\textit{Lossless}} & CopySpec & 16.31 & 172.8 & 77.24 & 300.4 & 80.35 & \textbf{96.3} & 67.20 & 334.2 & 60.28 & 225.9 \\
 & PICARD-style & 0.00 & 50.8 & 77.93 & 51.3 & 96.07 & 48.8 & 69.20 & 227.4 & 60.80 & 94.6 \\
\multirow{-2}{*}{\textit{Rule-based}} & Longest-Copy & 0.00 & -- & 12.41 & -- & 95.63 & -- & 0.00 & -- & 27.01 & -- \\
\rowcolor{gray!6} \textit{Learned} & CopySpan & 7.09 & 27.0 & 28.97 & 102.2 & 79.04 & 54.4 & 5.60 & 1335.3 & 30.18 & 379.7 \\
\rowcolor{green!8} \multicolumn{2}{l}{\textbf{\method\ (Ours)}} & \textbf{34.75} & 219.5 & \textbf{86.21} & \textbf{592.9} & \textbf{96.51} & 72.8 & \textbf{90.40} & \textbf{2483.2} & \textbf{76.97} & \textbf{842.1} \\
\midrule
% --- Ministral-8B ---
\rowcolor{gray!15} \multicolumn{12}{c}{\textit{\textbf{Ministral-8B}}} \\
\multicolumn{2}{l}{Base Model} & 19.15 & 60.0 & 41.38 & 62.1 & 96.07 & 63.4 & 42.80 & 60.9 & 49.85 & 61.6 \\
\rowcolor{gray!6} & LLMA & 19.15 & \textbf{220.3} & 41.38 & 469.4 & 96.07 & 94.7 & 42.80 & 53.4 & 49.85 & 209.5 \\
\rowcolor{gray!6} \multirow{-2}{*}{\textit{Lossless}} & CopySpec & 19.15 & 191.4 & 41.38 & 289.5 & 96.07 & 102.3 & 42.80 & 64.8 & 49.85 & 162.0 \\
 & PICARD-style & 0.00 & 60.0 & 42.07 & 60.4 & 87.77 & 59.2 & 40.80 & 82.1 & 42.66 & 65.4 \\
\multirow{-2}{*}{\textit{Rule-based}} & Longest-Copy & 0.00 & -- & 12.41 & -- & \textbf{96.51} & -- & 0.00 & -- & 27.23 & -- \\
\rowcolor{gray!6} \textit{Learned} & CopySpan & 2.13 & 58.9 & 19.31 & 415.7 & 93.01 & \textbf{514.3} & 54.00 & 410.9 & 42.11 & 349.9 \\
\rowcolor{green!8} \multicolumn{2}{l}{\textbf{\method\ (Ours)}} & \textbf{22.70} & 204.2 & \textbf{88.28} & \textbf{521.5} & \textbf{96.51} & 69.1 & \textbf{100.00} & \textbf{2609.6} & \textbf{76.87} & \textbf{851.1} \\
\bottomrule[1.5pt]
\end{tabular}
\end{adjustbox}
\label{tab:baseline_full}
\end{table*}

\subsection{Evaluation of Frontier LLMs}
\label{app:frontier}

Table~\ref{tab:api_models} compares three frontier LLMs with the open backbones under the same evaluation protocol, where the frontier LLMs are accessed through their official APIs. Results show that even the strongest frontier LLM falls well short of perfect exact copying.

\begin{table*}[h]
\caption{Exact match of frontier LLMs and open backbones with and without \method{} under the same evaluation protocol. Best results in each row are highlighted in \textbf{bold}.}
\centering
\renewcommand{\arraystretch}{1.2}
\setlength{\tabcolsep}{4.5pt}
\begin{adjustbox}{max width=\linewidth}
\begin{tabular}{l ccc ccc ccc ccc}
\toprule[1.5pt]
\multirow{2}{*}{\textbf{Domain}} & \multicolumn{3}{c}{\textbf{Frontier LLMs}} & \multicolumn{3}{c}{\textbf{Qwen3-4B}} & \multicolumn{3}{c}{\textbf{Qwen3-8B}} & \multicolumn{3}{c}{\textbf{Ministral-8B}} \\
\cmidrule(lr){2-4} \cmidrule(lr){5-7} \cmidrule(lr){8-10} \cmidrule(lr){11-13}
& \makecell{GPT-4.1\\-nano} & \makecell{Gemini-2.5\\-Flash-Lite} & \makecell{DeepSeek\\-V4-Pro}
& Base & \method & \method$_{\text{SFT}}$
& Base & \method & \method$_{\text{SFT}}$
& Base & \method & \method$_{\text{SFT}}$ \\
\midrule[1.5pt]
Contract   & 33.33 & 41.84 & 56.03 & 5.67 & 13.48 & 90.78 & 16.31 & 34.75 & 96.45 & 19.15 & 22.70 & \textbf{98.58} \\
Literature & 80.00 & 80.00 & 77.24 & 68.28 & 84.83 & 85.52 & 77.24 & 86.21 & \textbf{94.48} & 41.38 & 88.28 & 92.41 \\
Code       & 65.50 & 90.83 & 97.38 & 54.15 & 55.02 & 98.69 & 80.35 & 96.51 & 98.25 & 96.07 & 96.51 & \textbf{100.00} \\
Random     & 81.20 & 78.40 & 99.20 & 53.60 & 74.80 & 90.40 & 67.20 & 90.40 & 96.80 & 42.80 & \textbf{100.00} & \textbf{100.00} \\
\midrule
Avg        & 65.01 & 72.77 & 82.46 & 45.43 & 57.03 & 91.35 & 60.28 & 76.97 & 96.50 & 49.85 & 76.87 & \textbf{97.75} \\
\bottomrule[1.5pt]
\end{tabular}
\end{adjustbox}
\label{tab:api_models}
\end{table*}

\subsection{Ablation Study}
Table~\ref{tab:ablation_full} reports the full ablation results across all models and domains.
The results show that all three components contribute to the final performance: adaptive length proposal improves the speed--accuracy tradeoff, adaptive weighting improves verifier learning, and margin calibration is critical for reliable copy control.

\begin{table*}[h]
\caption{Full ablation results of \method. 
For \textit{w/o Length Proposal}, we replace the adaptive span proposal with fixed copy lengths $k \in \{8,16,32,64,128,256\}$ and report the average over the sweep.
Average is computed over all four domains. Best results are highlighted in \textbf{bold}.}
\centering
\renewcommand{\arraystretch}{1.12}
\begin{adjustbox}{max width=\linewidth}
\begin{tabular}{l l cc cc cc cc cc}
\toprule
\multirow{2}{*}{\textbf{Model}} & \multirow{2}{*}{\textbf{Variant}}
& \multicolumn{2}{c}{\textbf{Contract}}
& \multicolumn{2}{c}{\textbf{Literature}}
& \multicolumn{2}{c}{\textbf{Code}}
& \multicolumn{2}{c}{\textbf{Random Sequence}}
& \multicolumn{2}{c}{\textbf{Average}} \\
\cmidrule(lr){3-4}\cmidrule(lr){5-6}\cmidrule(lr){7-8}\cmidrule(lr){9-10}\cmidrule(lr){11-12}
&
& \textbf{EM} & \textbf{Tokens/s}
& \textbf{EM} & \textbf{Tokens/s}
& \textbf{EM} & \textbf{Tokens/s}
& \textbf{EM} & \textbf{Tokens/s}
& \textbf{EM} & \textbf{Tokens/s} \\
\midrule

\multirow{4}{*}{Qwen3-4B}
& \textbf{Full} & \textbf{13.48} & \textbf{238.02} & \textbf{84.83} & \textbf{729.42} & \textbf{55.02} & \textbf{132.28} & 74.80 & \textbf{2939.15} & \textbf{57.03} & \textbf{1009.72} \\
& w/o Length Proposal & 13.01 & 206.44 & 84.02 & 474.31 & 54.88 & 118.72 & \textbf{74.87} & 1010.51 & 56.69 & 452.49 \\
& w/o Adaptive Weighting & 12.77 & 223.38 & 68.28 & 606.32 & 53.28 & 123.41 & 52.80 & 1809.08 & 46.78 & 690.55 \\
& w/o Margin Calibration & 5.67 & 235.28 & 67.59 & 636.08 & 53.28 & 123.12 & 53.20 & 2198.09 & 44.94 & 798.14 \\
\midrule

\multirow{4}{*}{Qwen3-8B}
& \textbf{Full} & \textbf{34.75} & \textbf{219.54} & \textbf{86.21} & \textbf{592.85} & \textbf{96.51} & \textbf{72.79} & \textbf{90.40} & \textbf{2483.23} & \textbf{76.97} & \textbf{842.10} \\
& w/o Length Proposal & 29.79 & 173.06 & \textbf{86.21} & 379.71 & 94.47 & 69.78 & 86.80 & 797.19 & 74.32 & 354.94 \\
& w/o Adaptive Weighting & 29.79 & 194.64 & 76.55 & 530.84 & 80.35 & 71.24 & 68.00 & 1828.15 & 63.67 & 656.22 \\
& w/o Margin Calibration & 16.31 & 205.20 & 77.24 & 509.79 & 79.91 & 69.73 & 68.40 & 1895.75 & 60.47 & 670.12 \\

\midrule

\multirow{4}{*}{Ministral-8B}
& \textbf{Full} & \textbf{22.70} & \textbf{204.22} & \textbf{88.28} & \textbf{521.48} & \textbf{96.51} & \textbf{69.11} & \textbf{100.00} & \textbf{2609.59} & \textbf{76.87} & \textbf{851.10} \\
& w/o Length Proposal & 21.99 & 184.11 & 88.17 & 407.61 & \textbf{96.51} & 65.85 & 93.13 & 538.21 & 74.95 & 298.94 \\
& w/o Adaptive Weighting & 19.86 & 211.49 & 64.83 & 463.85 & 96.07 & 67.76 & 62.80 & 426.87 & 60.89 & 292.49 \\
& w/o Margin Calibration & 19.86 & 213.13 & 41.38 & 396.20 & 96.07 & 67.48 & 42.80 & 296.72 & 50.03 & 243.38 \\

\bottomrule
\end{tabular}
\end{adjustbox}
\label{tab:ablation_full}
\end{table*}

\subsection{Copy Dynamics Analysis}
\label{app:copy_analysis}

\begin{figure}[h]
    \centering
    \includegraphics[width=\textwidth]{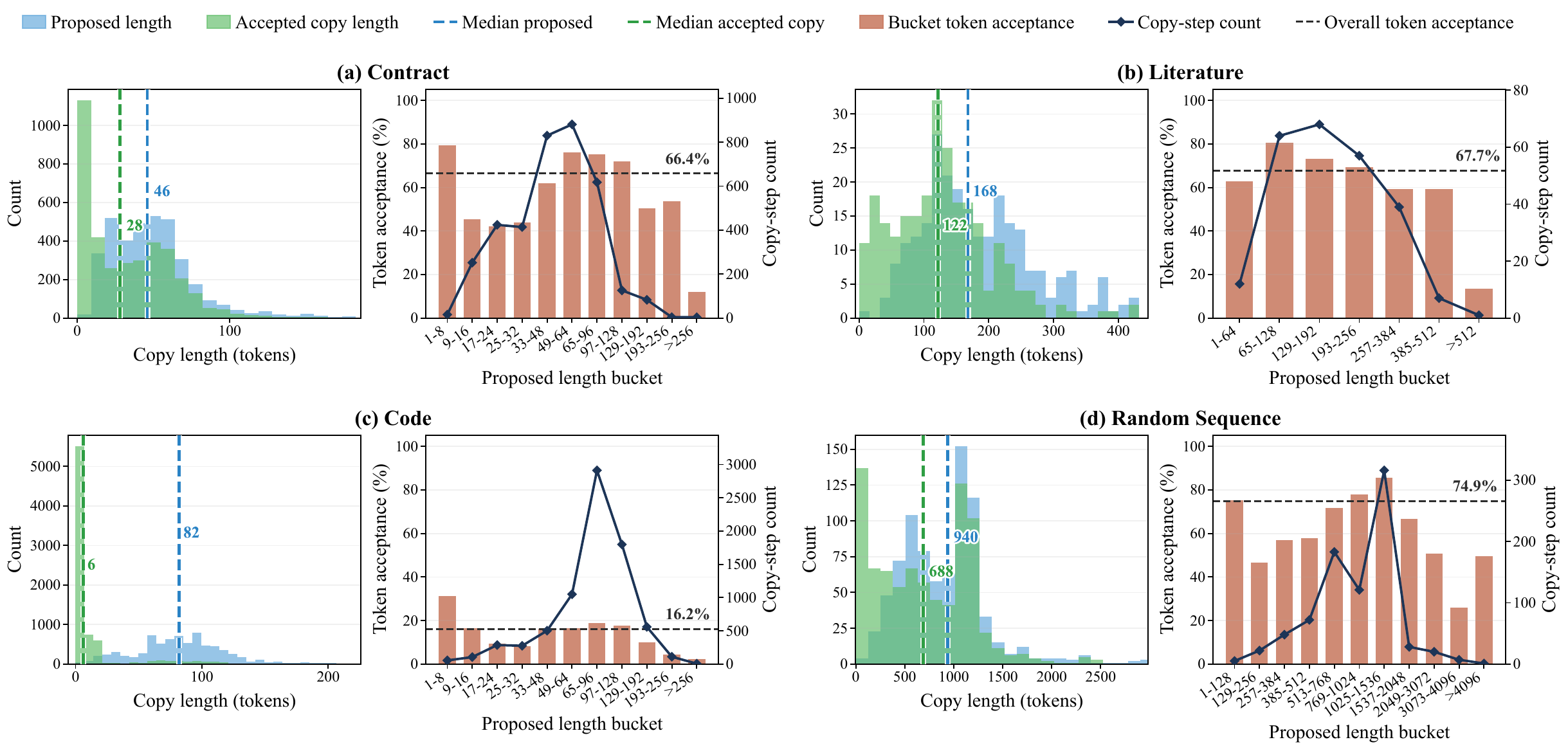}
        \caption{
        Copy dynamics for Qwen3-4B across four benchmark domains.
        For each domain, we show the distribution of proposed and accepted copy lengths (left), together with token acceptance by proposed-length bucket (right).
    }
        \label{fig:copy-dynamics-qwen-4b}
\end{figure}

\begin{figure}[h]
    \centering
    \includegraphics[width=\textwidth]{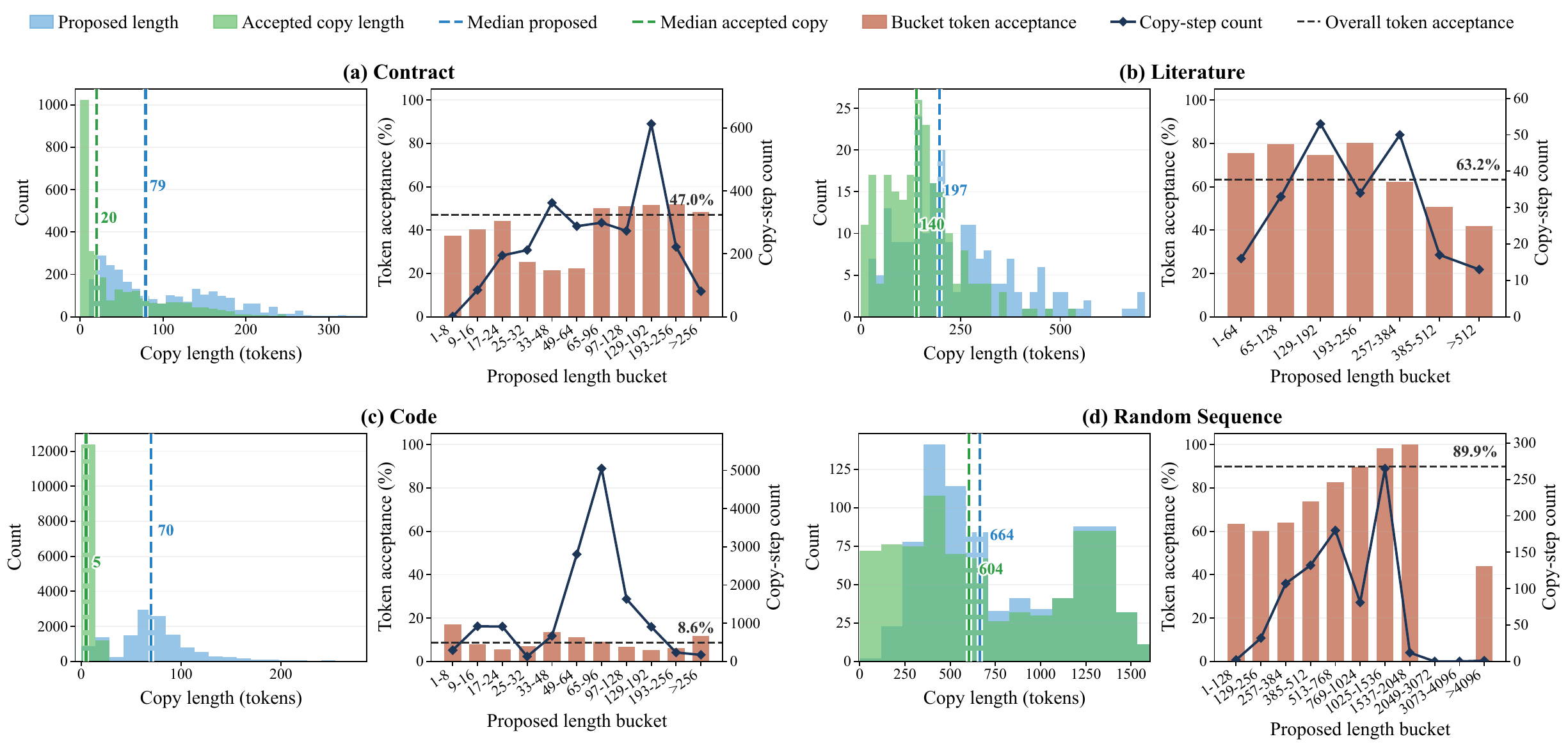}
        \caption{
        Copy dynamics for Qwen3-8B across four benchmark domains.
        For each domain, we show the distribution of proposed and accepted copy lengths (left), together with token acceptance by proposed-length bucket (right).
    }
    \label{fig:copy-dynamics-qwen-8b}
\end{figure}

\begin{figure}[h]
    \centering
    \includegraphics[width=\textwidth]{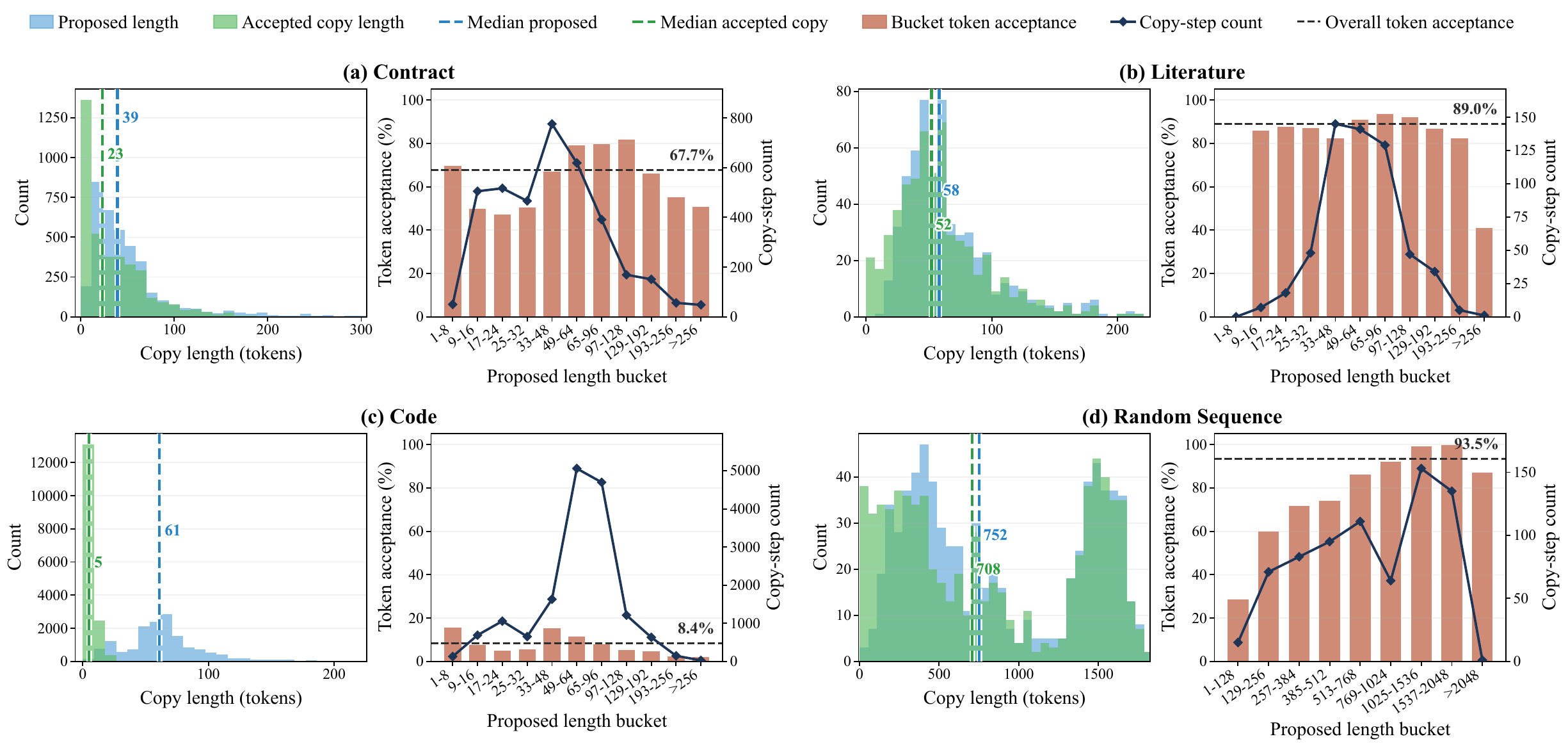}
        \caption{
        Copy dynamics for Ministral-8B across four benchmark domains.
        For each domain, we show the distribution of proposed and accepted copy lengths (left), together with token acceptance by proposed-length bucket (right).
    }
    \label{fig:copy-dynamics-ministral-8b}
\end{figure}

We provide the full results of the copy dynamics analysis, as shown in Figures~\ref{fig:copy-dynamics-qwen-4b}, \ref{fig:copy-dynamics-qwen-8b}, and~\ref{fig:copy-dynamics-ministral-8b}.
Across models and domains, the proposed lengths are often longer than the accepted lengths, showing that the length predictor provides tentative spans while the verifier determines the final copy boundary.
The bucket-level acceptance plots further show that longer proposed spans are not always fully accepted, highlighting the role of verification in preventing over-copy.

\subsection{Training Efficiency Analysis}
\label{app:training_res}

Table~\ref{tab:train_resource} reports the training resource usage of \method{} compared with SFT.
Across models and domains, \method{} substantially reduces peak GPU memory while maintaining comparable average training time per example.

\begin{table*}[h]
\caption{Training resource comparison. We report peak GPU memory during training (GiB) and average training time per example (s/item). Relative changes in parentheses are computed against SFT within each model and task. Lower is better for both metrics.}
\centering
\renewcommand{\arraystretch}{1.2}
\begin{adjustbox}{max width=\linewidth}
\begin{tabular}{l *{5}{cc}}
\toprule[1.5pt]
\multirow{2}{*}{\textbf{Method}} 
& \multicolumn{2}{c}{\textbf{Contract}} 
& \multicolumn{2}{c}{\textbf{Literature}} 
& \multicolumn{2}{c}{\textbf{Code}} 
& \multicolumn{2}{c}{\textbf{Random Sequence}} 
& \multicolumn{2}{c}{\textbf{Average}} \\
\cmidrule(lr){2-3} \cmidrule(lr){4-5} \cmidrule(lr){6-7} \cmidrule(lr){8-9} \cmidrule(lr){10-11}
& \textbf{Peak GiB} & \textbf{Train s/item}
& \textbf{Peak GiB} & \textbf{Train s/item}
& \textbf{Peak GiB} & \textbf{Train s/item}
& \textbf{Peak GiB} & \textbf{Train s/item}
& \textbf{Peak GiB} & \textbf{Train s/item} \\
\midrule[1.5pt]

% --- Model 1 ---
\rowcolor{gray!15} \multicolumn{11}{c}{\textit{\textbf{Qwen3-4B}}} \\
SFT & 27.62 & \textbf{1.316} & 21.28 & 2.610 & 18.43 & 2.749 & 41.67 & \textbf{1.064} & 27.25 & 1.935 \\
\rowcolor{green!3} \textbf{CopyGen} & \textbf{11.04} (-60.0\%) & 1.640 (+24.6\%) & \textbf{10.53} (-50.5\%) & \textbf{0.625} (-76.1\%) & \textbf{9.67} (-47.5\%) & 2.500 (-9.1\%) & \textbf{12.57} (-69.8\%) & 1.667 (+56.6\%) & \textbf{10.95} (-59.8\%) & \textbf{1.608} (-16.9\%) \\
\rowcolor{green!10} \textbf{CopyGen$_\text{SFT}$} & 11.08 (-59.9\%) & 1.800 (+36.8\%) & 10.56 (-50.4\%) & 1.667 (-36.1\%) & 9.72 (-47.3\%) & \textbf{2.000} (-27.2\%) & 12.63 (-69.7\%) & 1.429 (+34.3\%) & 11.00 (-59.6\%) & 1.724 (-10.9\%) \\
\midrule

% --- Model 2 ---
\rowcolor{gray!15} \multicolumn{11}{c}{\textit{\textbf{Qwen3-8B}}} \\
SFT & 36.31 & \textbf{1.897} & 29.40 & 1.795 & 26.45 & 1.967 & 50.05 & \textbf{0.729} & 35.55 & 1.597 \\
\rowcolor{green!3} \textbf{CopyGen} & \textbf{19.19} (-47.1\%) & 2.370 (+24.9\%) & \textbf{18.47} (-37.2\%) & \textbf{1.250} (-30.4\%) & \textbf{17.67} (-33.2\%) & 1.667 (-15.3\%) & \textbf{20.70} (-58.6\%) & 1.111 (+52.4\%) & \textbf{19.01} (-46.5\%) & 1.600 (+0.2\%) \\
\rowcolor{green!10} \textbf{CopyGen$_\text{SFT}$} & 19.24 (-47.0\%) & 2.680 (+41.2\%) & 18.49 (-37.1\%) & \textbf{1.250} (-30.4\%) & 17.74 (-32.9\%) & \textbf{1.429} (-27.4\%) & 20.76 (-58.5\%) & 1.000 (+37.2\%) & 19.06 (-46.4\%) & \textbf{1.590} (-0.5\%) \\
\midrule

% --- Model 3 ---
\rowcolor{gray!15} \multicolumn{11}{c}{\textit{\textbf{Ministral-8B}}} \\
SFT & 32.02 & \textbf{1.748} & 26.74 & 2.060 & 24.92 & 2.029 & 46.34 & \textbf{0.718} & 32.51 & 1.639 \\
\rowcolor{green!3} \textbf{CopyGen} & \textbf{19.01} (-40.6\%) & 2.390 (+36.7\%) & \textbf{17.87} (-33.2\%) & 1.429 (-30.7\%) & \textbf{17.44} (-30.0\%) & 1.667 (-17.9\%) & 20.27 (-56.3\%) & 1.111 (+54.7\%) & \textbf{18.65} (-42.6\%) & 1.649 (+0.6\%) \\
\rowcolor{green!10} \textbf{CopyGen$_\text{SFT}$} & 19.07 (-40.4\%) & 2.720 (+55.6\%) & 17.91 (-33.0\%) & \textbf{1.250} (-39.3\%) & 17.52 (-29.7\%) & \textbf{1.429} (-29.6\%) & \textbf{20.22} (-56.4\%) & 1.000 (+39.3\%) & 18.68 (-42.5\%) & \textbf{1.600} (-2.4\%) \\
\bottomrule[1.5pt]
\end{tabular}
\end{adjustbox}
\label{tab:train_resource}
\end{table*}

\section{Licenses and Asset Usage}
\label{app:licenses}

We document the external assets used in this work, including model backbones, data sources, and code repositories, together with their licenses or terms of use.

\textbf{Model Backbones.}
We evaluate \texttt{Qwen3-4B} and \texttt{Qwen3-8B}, released by Qwen under the Apache 2.0 License and available at \url{https://huggingface.co/Qwen/Qwen3-4B} and \url{https://huggingface.co/Qwen/Qwen3-8B}. We also evaluate \texttt{Ministral-8B-Instruct}, released by Mistral AI under the Mistral Research License and available at \url{https://huggingface.co/mistralai/Ministral-8B-Instruct-2410}. All model backbones are used for research evaluation in accordance with their corresponding licenses and model cards.

\textbf{Contract Data.}
The contract task is constructed from publicly accessible SEC EDGAR filings, available through the SEC EDGAR system at \url{https://www.sec.gov/edgar/search/}. We use these filings for research purposes and follow the SEC access policies, including fair-access requirements for automated retrieval.

\textbf{Scientific Literature Data.}
The scientific-literature task is constructed from full-text articles in the PubMed Central Open Access subset, available at \url{https://pmc.ncbi.nlm.nih.gov/}. Since PMC Open Access articles are distributed under article-level licenses, we follow the license terms associated with each article.

\textbf{Code Data.}
The code task is constructed from publicly available open-source Python repositories, including \texttt{FastAPI}, \texttt{SQLModel}, and \texttt{Starlette}. \texttt{FastAPI} and \texttt{SQLModel} are released under the MIT License, and \texttt{Starlette} is released under the BSD 3-Clause License. These repositories are available at \url{https://github.com/fastapi/fastapi}, \url{https://github.com/fastapi/sqlmodel}, and \url{https://github.com/Kludex/starlette}.

\textbf{Random Sequences.}
The random-sequence task is generated by us. Although the strings are formatted to resemble credential-like or encoded configuration values, all such strings are synthetic and do not correspond to real credentials, secrets, or private keys.

All external assets are credited to their original creators or maintainers, with source URLs and license information provided above where applicable. 
We use these assets only for research and evaluation purposes and in compliance with their corresponding licenses, model cards, access policies, or terms of use. We do not use proprietary, private, or restricted-access data in this study.

\clearpage
\section{Prompt Templates}
\label{app:prompts}

\begin{tcolorbox}[
    enhanced,
    adjusted title,
    breakable,
    float*=h,  
    width=\textwidth,  
    title={\textbf{\textsf{Code}}},
    coltitle=black,
    colbacktitle=boxtitle,
    fonttitle=\small\sffamily,
    attach boxed title to top left={xshift=10pt, yshift*=-12pt},
    boxed title style={
        colback=boxtitle,
        colframe=boxframe,
        boxrule=0.5pt,
        arc=2pt,
        outer arc=2pt,
    },
    colframe=boxframe,
    colback=boxback,
    boxrule=0.8pt,
    left=6pt, right=6pt, top=12pt, bottom=6pt,
    arc=3pt,
    drop shadow=black!15!white, 
]

\begin{lstlisting}[
    basicstyle=\ttfamily\footnotesize,
    breaklines=true,
    columns=fullflexible,
    mathescape=true,
    frame=none 
]
System:
You are a precise long-context text assistant. Follow the user's instruction over the provided source text. When the instruction asks for exact copying or quoting, preserve the source wording exactly. Output only the requested final answer. Do not add explanations, notes, markdown fences, or any extra content.

User:
Instruction:

In the Python function `<function_name>`, briefly explain what the anchored block is used for, then quote exactly the block beginning with `<anchor_line>`, ending at the matching closing line for that block. Preserve every line, indentation level, quote, comma, bracket, and any trailing comma on the closing line.

Source text:

[FILE]
<file_name>

[LINES]
<start_line>-<end_line>

````python
<local_python_context>
Constraint:
Return only the requested answer. Nothing else.
\end{lstlisting}
\end{tcolorbox}

\begin{tcolorbox}[
    enhanced,
    adjusted title,
    breakable,
    float*=h,
    width=\textwidth,
    title={\textbf{\textsf{Literature}}},
    coltitle=black,
    colbacktitle=boxtitle,
    fonttitle=\small\sffamily,
    attach boxed title to top left={xshift=10pt, yshift*=-12pt},
    boxed title style={
        colback=boxtitle,
        colframe=boxframe,
        boxrule=0.5pt,
        arc=2pt,
        outer arc=2pt,
    },
    colframe=boxframe,
    colback=boxback,
    boxrule=0.8pt,
    left=6pt, right=6pt, top=12pt, bottom=6pt,
    arc=3pt,
    drop shadow=black!15!white,
]
\begin{lstlisting}[
    basicstyle=\ttfamily\footnotesize,
    breaklines=true,
    columns=fullflexible,
    mathescape=true,
    frame=none
]
System:
You are a precise long-context text assistant. Follow the user's instruction over the provided source text. When the instruction asks for exact copying or quoting, preserve the source wording exactly. Output only the requested final answer. Do not add explanations, notes, markdown fences, or any extra content.

User:
Instruction:

In the paper "<paper_title>", return the exact content from "<section_title>" that begins with "<anchor_text>" and provides the requested scientific detail. Output only the copied text.

Source text:

<paper_title>

<section_context>

Constraint:
Return only the requested answer. Nothing else.
\end{lstlisting}
\end{tcolorbox}

\begin{tcolorbox}[
    enhanced,
    adjusted title,
    breakable,
    float*=h,
    width=\textwidth,
    title={\textbf{\textsf{Contract}}},
    coltitle=black,
    colbacktitle=boxtitle,
    fonttitle=\small\sffamily,
    attach boxed title to top left={xshift=10pt, yshift*=-12pt},
    boxed title style={
        colback=boxtitle,
        colframe=boxframe,
        boxrule=0.5pt,
        arc=2pt,
        outer arc=2pt,
    },
    colframe=boxframe,
    colback=boxback,
    boxrule=0.8pt,
    left=6pt, right=6pt, top=12pt, bottom=6pt,
    arc=3pt,
    drop shadow=black!15!white,
]
\begin{lstlisting}[
    basicstyle=\ttfamily\footnotesize,
    breaklines=true,
    columns=fullflexible,
    mathescape=true,
    frame=none
]
System:
You are a precise long-context text assistant. Follow the user's instruction over the provided source text. When the instruction asks for exact copying or quoting, preserve the source wording exactly. Output only the requested final answer. Do not add explanations, notes, markdown fences, or any extra content.

User:
Instruction:

Extract the full text of the following clauses exactly as written in the article below:
1. the clause headed "<clause_heading_1>".
2. the clause headed "<clause_heading_2>".
...

Return a JSON object with keys "<clause_key_1>", "<clause_key_2>", ... . Include each section heading and preserve the original wording, line breaks, and indentation. Output only the JSON object.

Source text:

<contract_article_context>

Constraint:
Return only the requested answer. Nothing else.
\end{lstlisting}
\end{tcolorbox}

\begin{tcolorbox}[
    enhanced,
    adjusted title,
    breakable,
    float*=h,
    width=\textwidth,
    title={\textbf{\textsf{Random Sequence}}},
    coltitle=black,
    colbacktitle=boxtitle,
    fonttitle=\small\sffamily,
    attach boxed title to top left={xshift=10pt, yshift*=-12pt},
    boxed title style={
        colback=boxtitle,
        colframe=boxframe,
        boxrule=0.5pt,
        arc=2pt,
        outer arc=2pt,
    },
    colframe=boxframe,
    colback=boxback,
    boxrule=0.8pt,
    left=6pt, right=6pt, top=12pt, bottom=6pt,
    arc=3pt,
    drop shadow=black!15!white,
]
\begin{lstlisting}[
    basicstyle=\ttfamily\footnotesize,
    breaklines=true,
    columns=fullflexible,
    mathescape=true,
    frame=none
]
System:
You are a precise long-context text assistant. Follow the user's instruction over the provided source text. When the instruction asks for exact copying or quoting, preserve the source wording exactly. Output only the requested final answer. Do not add explanations, notes, markdown fences, or any extra content.

User:
Instruction:

Return only the exact string content for the given text. Preserve line breaks exactly as written. Output only the copied content.

Source text:

<SOURCE TEXT>

Constraint:
Return only the requested answer. Nothing else.
\end{lstlisting}
\end{tcolorbox}

\clearpage
\section{Task Examples}

Here, we show one example for each task type to make the task format concrete.

% warm yellow task-example colors
\definecolor{taskyellowback}{HTML}{FFFDF3}
\definecolor{taskyellowtitle}{HTML}{F9E6A6}
\definecolor{taskyellowframe}{HTML}{D8B86A}
\definecolor{taskyellowlabel}{HTML}{8A650F}

\begin{tcolorbox}[
    enhanced,
    adjusted title,
    breakable,
    float*=h,
    width=\textwidth,
    title={\textbf{\textsf{Contract}}},
    coltitle=black,
    colbacktitle=taskyellowtitle,
    fonttitle=\small\sffamily,
    attach boxed title to top left={xshift=10pt, yshift*=-12pt},
    boxed title style={
        colback=taskyellowtitle,
        colframe=taskyellowframe,
        boxrule=0.5pt,
        arc=2pt,
        outer arc=2pt,
    },
    colframe=taskyellowframe,
    colback=taskyellowback,
    boxrule=0.8pt,
    left=7pt, right=7pt, top=13pt, bottom=7pt,
    arc=3pt,
    drop shadow=black!10!white,
]
\scriptsize
\textbf{\textcolor{taskyellowlabel}{Input.}}
Extract the full text of the following clauses exactly as written in the article below:
1. the clause headed ``CHOICE OF LAW.''
2. the clause headed ``CONSENT TO JURISDICTION.''
3. the clause headed ``Entire Agreement.''
4. the clause headed ``Survival of Representations.''

Return a JSON object with keys \texttt{governing\_law\_clause}, \texttt{consent\_to\_jurisdiction\_clause}, \texttt{entire\_agreement\_clause}, \texttt{survival\_clause}. Include each section heading and preserve the original wording, line breaks, and indentation. Output only the JSON object.

\vspace{2pt}
\textbf{\textcolor{taskyellowlabel}{Source excerpt.}}
\begin{lstlisting}[
    basicstyle=\ttfamily\tiny,
    breaklines=true,
    columns=fullflexible,
    frame=none,
    aboveskip=2pt,
    belowskip=2pt
]
ARTICLE X
GENERAL PROVISIONS

10.1. Survival of Representations.
All representations and warranties made hereunder and in any other Loan Document or other document delivered pursuant hereto or thereto or in 201380721_5

connection herewith or therewith shall survive the execution and delivery hereof and thereof. Such representations and warranties have been or will be relied upon by the Administrative Agent and each Bank ...

...

10.4. Entire Agreement.
This Agreement, together with the other Loan Documents, comprises the complete and integrated agreement of the parties on the subject matter hereof and thereof and supersedes all prior agreements, written or oral, on such subject matter ...

...

10.12. CHOICE OF LAW.
THE LOAN DOCUMENTS (OTHER THAN THOSE CONTAINING A CONTRARY EXPRESS CHOICE OF LAW PROVISION) SHALL BE CONSTRUED IN ACCORDANCE WITH THE INTERNAL LAWS ...

10.13. CONSENT TO JURISDICTION.
ANY LEGAL ACTION, LITIGATION OR PROCEEDING WITH RESPECT TO THIS AGREEMENT OR ANY OTHER LOAN DOCUMENT MAY BE BROUGHT IN THE COURTS OF THE STATE OF NEW YORK ...

10.14. WAIVER OF JURY TRIAL.
EACH LOAN PARTY, THE ADMINISTRATIVE AGENT AND EACH BANK HEREBY WAIVE TRIAL BY JURY ...
\end{lstlisting}

\vspace{2pt}
\textbf{\textcolor{taskyellowlabel}{Output.}}
\begin{lstlisting}[
    basicstyle=\ttfamily\tiny,
    breaklines=true,
    columns=fullflexible,
    frame=none,
    aboveskip=2pt,
    belowskip=2pt
]
{
  "governing_law_clause": "10.12. CHOICE OF LAW.\nTHE LOAN DOCUMENTS (OTHER THAN THOSE CONTAINING A CONTRARY EXPRESS CHOICE OF LAW PROVISION) SHALL BE CONSTRUED IN ACCORDANCE WITH THE INTERNAL LAWS (AND NOT THE LAW OF CONFLICTS) OF THE STATE OF NEW YORK BUT GIVING EFFECT TO FEDERAL LAWS APPLICABLE TO NATIONAL BANKS.",
  "consent_to_jurisdiction_clause": "10.13. CONSENT TO JURISDICTION.\nANY LEGAL ACTION, LITIGATION OR PROCEEDING WITH RESPECT TO THIS AGREEMENT OR ANY OTHER LOAN DOCUMENT MAY BE BROUGHT IN THE COURTS OF THE STATE OF NEW YORK SITTING IN NEW YORK COUNTY OR OF THE UNITED STATES DISTRICT COURT OF THE SOUTHERN DISTRICT OF NEW YORK SITTING IN NEW YORK COUNTY, AND ANY APPELLATE COURT FROM ANY THEREOF, AND EACH OF THE PARTIES HERETO IRREVOCABLY AND UNCONDITIONALLY SUBMITS TO THE JURISDICTION OF SUCH COURTS AND AGREES THAT ALL CLAIMS IN RESPECT OF ANY SUCH ACTION, LITIGATION OR PROCEEDING MAY BE HEARD AND DETERMINED IN SUCH NEW YORK STATE COURT OR, TO THE FULLEST EXTENT PERMITTED BY APPLICABLE LAW, IN SUCH FEDERAL COURT. EACH OF THE PARTIES HERETO HEREBY IRREVOCABLY WAIVES ANY OBJECTION IT MAY NOW OR HEREAFTER HAVE AS TO THE VENUE OF ANY SUCH SUIT, ACTION OR PROCEEDING BROUGHT IN SUCH A COURT OR THAT SUCH COURT IS AN INCONVENIENT FORUM.",
  "entire_agreement_clause": "10.4. Entire Agreement.\nThis Agreement, together with the other Loan Documents, comprises the complete and integrated agreement of the parties on the subject matter hereof and thereof and supersedes all prior agreements, written or oral, on such subject matter. In the event of any conflict between the provisions of this Agreement and those of any other Loan Document, the provisions of this Agreement shall control; provided, that the inclusion of supplemental rights or remedies in favor of the Administrative Agent or the Banks in any other Loan Document shall not be deemed a conflict with this Agreement. Each Loan Document was drafted with the joint participation of the respective parties thereto and shall be construed neither against nor in favor of any party, but rather in accordance with the fair meaning thereof.",
  "survival_clause": "10.1. Survival of Representations.\nAll representations and warranties made hereunder and in any other Loan Document or other document delivered pursuant hereto or thereto or in 201380721_5\n\nconnection herewith or therewith shall survive the execution and delivery hereof and thereof. Such representations and warranties have been or will be relied upon by the Administrative Agent and each Bank, regardless of any investigation made by the Administrative Agent or any Bank or on their behalf and notwithstanding that the Administrative Agent or any Bank may have had notice or knowledge of any Default or Unmatured Default at the time of any Loan or Commitment, and shall continue in full force and effect as long as any Loan or any other Obligation shall remain unpaid or unsatisfied."
}
\end{lstlisting}
\end{tcolorbox}

\begin{tcolorbox}[
    enhanced,
    breakable,
    float*=h,
    width=\textwidth,
    title={\textbf{\textsf{Literature}}},
    coltitle=black,
    colbacktitle=taskyellowtitle,
    fonttitle=\small\sffamily,
    attach boxed title to top left={xshift=10pt, yshift*=-12pt},
    boxed title style={
        colback=taskyellowtitle,
        colframe=taskyellowframe,
        boxrule=0.5pt,
        arc=2pt,
        outer arc=2pt,
    },
    colframe=taskyellowframe,
    colback=taskyellowback,
    boxrule=0.8pt,
    left=8pt, right=8pt, top=13pt, bottom=8pt,
    arc=3pt,
    drop shadow=black!12!white,
]
\small
\textbf{\textcolor{taskyellowlabel}{Input.}}
From the paper \emph{Inhibition of cell growth by EGR-1 in human primary cultures from malignant glioma}, extract the exact content from the subsection headed \emph{DNA Sequencing of p53 Exons 5--9} that contains the sequencing or genotyping procedure and begins with ``For analysis of p53 mutations, previously described primer sets were used for PCR amplification from high molecular weight DNA of three DNA fragments corresponding to p53 exons 5 and 6, exon 7, and exons 8 and 9, respectively.'' Output only the copied text.

\vspace{4pt}
\textbf{\textcolor{taskyellowlabel}{Source excerpt.}}
Densitometric analysis of Western blots was performed on a Macintosh G3/233 computer using the public domain NIH Image software. The EGR-1 and fibronectin chemiluminescence signals were normalized with respect to actin content of each sample.

\centerline{\textcolor{taskyellowlabel}{\(\cdots\)}}

\emph{DNA Sequencing of p53 Exons 5--9.}
For analysis of p53 mutations, previously described primer sets were used for PCR amplification from high molecular weight DNA of three DNA fragments corresponding to p53 exons 5 and 6, exon 7, and exons 8 and 9, respectively [33]. Direct sequencing of the specific DNA fragments was performed using a Big Dye terminator DNA sequencing kit with the ABI PRISM 377 DNA Sequencer (PE Applied Biosystems Inc., Foster City, CA, USA), according to the manufacturer's instructions. Results were analyzed using the ABI sequencing analysis software.

\centerline{\textcolor{taskyellowlabel}{\(\cdots\)}}

\emph{Analysis of allele dosage for MDM2 and loss of heterozygosity for p16/ARF.}
Southern blotting was used for monitoring MDM2 gene amplification and loss of heterozygosity at the p16INK4a/ARF locus.

\vspace{4pt}
\textbf{\textcolor{taskyellowlabel}{Output.}}
For analysis of p53 mutations, previously described primer sets were used for PCR amplification from high molecular weight DNA of three DNA fragments corresponding to p53 exons 5 and 6, exon 7, and exons 8 and 9, respectively. Direct sequencing of the specific DNA fragments was performed using a Big Dye terminator DNA sequencing kit with the ABI PRISM 377 DNA Sequencer (PE Applied Biosystems Inc., Foster City, CA, USA), according to the manufacturer's instructions. Results were analyzed using the ABI sequencing analysis software.
\end{tcolorbox}

\begin{tcolorbox}[
    enhanced,
    adjusted title,
    breakable,
    float*=h,
    width=\textwidth,
    title={\textbf{\textsf{Code}}},
    coltitle=black,
    colbacktitle=taskyellowtitle,
    fonttitle=\small\sffamily,
    attach boxed title to top left={xshift=10pt, yshift*=-12pt},
    boxed title style={
        colback=taskyellowtitle,
        colframe=taskyellowframe,
        boxrule=0.5pt,
        arc=2pt,
        outer arc=2pt,
    },
    colframe=taskyellowframe,
    colback=taskyellowback,
    boxrule=0.8pt,
    left=7pt, right=7pt, top=13pt, bottom=7pt,
    arc=3pt,
    drop shadow=black!10!white,
]
\scriptsize
\textbf{\textcolor{taskyellowlabel}{Input.}}
In the Python function \texttt{test\_foo\_no\_needy}, briefly explain what the anchored block is used for, then quote exactly the block beginning with \texttt{"detail": [} for key \texttt{detail}, ending at the matching closing line for that block.

\vspace{2pt}
\textbf{\textcolor{taskyellowlabel}{Source excerpt.}}
\begin{lstlisting}[
    basicstyle=\ttfamily\tiny,
    breaklines=true,
    columns=fullflexible,
    frame=none,
    aboveskip=2pt,
    belowskip=2pt
]
def test_foo_no_needy(client: TestClient):
    response = client.get("/items/foo?skip=a&limit=b")
    assert response.status_code == 422
    assert response.json() == {
        "detail": [
            {
                "type": "missing",
                "loc": ["query", "needy"],
                "msg": "Field required",
                "input": None,
            },
            {
                "type": "int_parsing",
                "loc": ["query", "skip"],
                "msg": "Input should be a valid integer, unable to parse string as an integer",
                "input": "a",
            },
            {
                "type": "int_parsing",
                "loc": ["query", "limit"],
                "msg": "Input should be a valid integer, unable to parse string as an integer",
                "input": "b",
            },
        ]
    }
\end{lstlisting}

\vspace{2pt}
\textbf{\textcolor{taskyellowlabel}{Output.}}
\begin{lstlisting}[
    basicstyle=\ttfamily\tiny,
    breaklines=true,
    columns=fullflexible,
    frame=none,
    aboveskip=2pt,
    belowskip=2pt
]
Explanation:
This block specifies the expected validation errors for a request with a missing required query parameter and two non-integer query values.

Quoted block:
````python
        "detail": [
            {
                "type": "missing",
                "loc": ["query", "needy"],
                "msg": "Field required",
                "input": None,
            },
            {
                "type": "int_parsing",
                "loc": ["query", "skip"],
                "msg": "Input should be a valid integer, unable to parse string as an integer",
                "input": "a",
            },
            {
                "type": "int_parsing",
                "loc": ["query", "limit"],
                "msg": "Input should be a valid integer, unable to parse string as an integer",
                "input": "b",
            },
        ]
\end{lstlisting}
\end{tcolorbox}

\begin{tcolorbox}[
    enhanced,
    adjusted title,
    breakable,
    float*=h,
    width=\textwidth,
    title={\textbf{\textsf{Random Sequence}}},
    coltitle=black,
    colbacktitle=taskyellowtitle,
    fonttitle=\small\sffamily,
    attach boxed title to top left={xshift=10pt, yshift*=-12pt},
    boxed title style={
        colback=taskyellowtitle,
        colframe=taskyellowframe,
        boxrule=0.5pt,
        arc=2pt,
        outer arc=2pt,
    },
    colframe=taskyellowframe,
    colback=taskyellowback,
    boxrule=0.8pt,
    left=7pt, right=7pt, top=13pt, bottom=7pt,
    arc=3pt,
    drop shadow=black!10!white,
]
\small
The random-sequence task uses synthetic key-like strings where every character and line break must be preserved. 

\begin{lstlisting}[
    basicstyle=\ttfamily\tiny,
    breaklines=true,
    columns=fullflexible,
    frame=none,
    aboveskip=3pt,
    belowskip=2pt
]
-----BEGIN OPENSSH PRIVATE KEY-----
b3BlbnNzaC1rZXktdjEAQdqHI9+PedkeGKrzeay8vXCutJJT7hJ9U00gWwS3adFRpogbQy
+SMlY9zhB9CxS8qRI6QTJMgzmdGplVr9cJRsSLEm3C61btnf2uyaajGLiesOFHRDsRMLP6
dDm4SC1tMWgPHyubKvqWwbf9F5RlwDvBPmPLRkC3jWCcEnVNHwO3AI0UG9N0qPv0ntLEGr
GmzYmikeJax2eh72D+Iv8n+iXmDiOnu9EXLjCncRyeF9Jxb14poJ16YIDh9mtS6g94sw5n
YzYCdulqqLKrPKkyHOAsv2quMkHrvARiOuHd9aaeoHipcSHjNJE6GC/cvSJTqZMHDKcPxw
d3RDCK6c+eLDRcyHfGvYYMnOPEYEPfYjGMQZuUTQK8As9scjSnF96EiYcMlUi12S4UULfz
...
VHtWPskfnjud3jKIdCuT6byHo3dSzzHjidub4JNtTrT6NFb21KapuDg2v5DYeV/eTRO1bH
P1N3qpn+yu5RYEtNtWK0TYyY6maGNlBjxeF0zVyDMnRKu03a93yA23vKhSTSnSvCyghwxo
F4oS9tSsvkg5ItCHlEVi4te8y9jP+2XpMeSvmvXWaRC/t0GzkPs6mSkMm+r14pM5r4vKw0
6bd883
-----END OPENSSH PRIVATE KEY-----
\end{lstlisting}
\end{tcolorbox}

\clearpage
\section*{NeurIPS Paper Checklist}

\begin{enumerate}

\item {\bf Claims}
    \item[] Question: Do the main claims made in the abstract and introduction accurately reflect the paper's contributions and scope?
    \item[] Answer: \answerYes{}
    \item[] Justification: The abstract and Introduction summarize the main claims and scope of our work, and the empirical results in Section~\ref{sec:experiment} support these claims.
    \item[] Guidelines:
    \begin{itemize}
        \item The answer \answerNA{} means that the abstract and introduction do not include the claims made in the paper.
        \item The abstract and/or introduction should clearly state the claims made, including the contributions made in the paper and important assumptions and limitations. A \answerNo{} or \answerNA{} answer to this question will not be perceived well by the reviewers. 
        \item The claims made should match theoretical and experimental results, and reflect how much the results can be expected to generalize to other settings. 
        \item It is fine to include aspirational goals as motivation as long as it is clear that these goals are not attained by the paper. 
    \end{itemize}

\item {\bf Limitations}
    \item[] Question: Does the paper discuss the limitations of the work performed by the authors?
    \item[] Answer: \answerYes{}
    \item[] Justification: Appendix~\ref{app:limitations} discusses the main limitations of our work.
    \item[] Guidelines:
    \begin{itemize}
        \item The answer \answerNA{} means that the paper has no limitation while the answer \answerNo{} means that the paper has limitations, but those are not discussed in the paper. 
        \item The authors are encouraged to create a separate ``Limitations'' section in their paper.
        \item The paper should point out any strong assumptions and how robust the results are to violations of these assumptions (e.g., independence assumptions, noiseless settings, model well-specification, asymptotic approximations only holding locally). The authors should reflect on how these assumptions might be violated in practice and what the implications would be.
        \item The authors should reflect on the scope of the claims made, e.g., if the approach was only tested on a few datasets or with a few runs. In general, empirical results often depend on implicit assumptions, which should be articulated.
        \item The authors should reflect on the factors that influence the performance of the approach. For example, a facial recognition algorithm may perform poorly when image resolution is low or images are taken in low lighting. Or a speech-to-text system might not be used reliably to provide closed captions for online lectures because it fails to handle technical jargon.
        \item The authors should discuss the computational efficiency of the proposed algorithms and how they scale with dataset size.
        \item If applicable, the authors should discuss possible limitations of their approach to address problems of privacy and fairness.
        \item While the authors might fear that complete honesty about limitations might be used by reviewers as grounds for rejection, a worse outcome might be that reviewers discover limitations that aren't acknowledged in the paper. The authors should use their best judgment and recognize that individual actions in favor of transparency play an important role in developing norms that preserve the integrity of the community. Reviewers will be specifically instructed to not penalize honesty concerning limitations.
    \end{itemize}

\item {\bf Theory assumptions and proofs}
    \item[] Question: For each theoretical result, does the paper provide the full set of assumptions and a complete (and correct) proof?
    \item[] Answer: \answerNA{} % Replace by \answerYes{}, \answerNo{}, or \answerNA{}.
    \item[] Justification: This paper does not include theoretical results or formal proofs.
    \item[] Guidelines:
    \begin{itemize}
        \item The answer \answerNA{} means that the paper does not include theoretical results. 
        \item All the theorems, formulas, and proofs in the paper should be numbered and cross-referenced.
        \item All assumptions should be clearly stated or referenced in the statement of any theorems.
        \item The proofs can either appear in the main paper or the supplemental material, but if they appear in the supplemental material, the authors are encouraged to provide a short proof sketch to provide intuition. 
        \item Inversely, any informal proof provided in the core of the paper should be complemented by formal proofs provided in appendix or supplemental material.
        \item Theorems and Lemmas that the proof relies upon should be properly referenced. 
    \end{itemize}

    \item {\bf Experimental result reproducibility}
    \item[] Question: Does the paper fully disclose all the information needed to reproduce the main experimental results of the paper to the extent that it affects the main claims and/or conclusions of the paper (regardless of whether the code and data are provided or not)?
    \item[] Answer: \answerYes{} % Replace by \answerYes{}, \answerNo{}, or \answerNA{}.
    \item[] Justification: Sections~\ref{sec:method}--\ref{sec:experiment} and appendix \ref{app:impl_details} provide the method, benchmark construction, implementation details, and evaluation setup needed to reproduce the main results.
    \item[] Guidelines:
    \begin{itemize}
        \item The answer \answerNA{} means that the paper does not include experiments.
        \item If the paper includes experiments, a \answerNo{} answer to this question will not be perceived well by the reviewers: Making the paper reproducible is important, regardless of whether the code and data are provided or not.
        \item If the contribution is a dataset and\slash or model, the authors should describe the steps taken to make their results reproducible or verifiable. 
        \item Depending on the contribution, reproducibility can be accomplished in various ways. For example, if the contribution is a novel architecture, describing the architecture fully might suffice, or if the contribution is a specific model and empirical evaluation, it may be necessary to either make it possible for others to replicate the model with the same dataset, or provide access to the model. In general. releasing code and data is often one good way to accomplish this, but reproducibility can also be provided via detailed instructions for how to replicate the results, access to a hosted model (e.g., in the case of a large language model), releasing of a model checkpoint, or other means that are appropriate to the research performed.
        \item While NeurIPS does not require releasing code, the conference does require all submissions to provide some reasonable avenue for reproducibility, which may depend on the nature of the contribution. For example
        \begin{enumerate}
            \item If the contribution is primarily a new algorithm, the paper should make it clear how to reproduce that algorithm.
            \item If the contribution is primarily a new model architecture, the paper should describe the architecture clearly and fully.
            \item If the contribution is a new model (e.g., a large language model), then there should either be a way to access this model for reproducing the results or a way to reproduce the model (e.g., with an open-source dataset or instructions for how to construct the dataset).
            \item We recognize that reproducibility may be tricky in some cases, in which case authors are welcome to describe the particular way they provide for reproducibility. In the case of closed-source models, it may be that access to the model is limited in some way (e.g., to registered users), but it should be possible for other researchers to have some path to reproducing or verifying the results.
        \end{enumerate}
    \end{itemize}

\item {\bf Open access to data and code}
    \item[] Question: Does the paper provide open access to the data and code, with sufficient instructions to faithfully reproduce the main experimental results, as described in supplemental material?
    \item[] Answer: \answerYes{} % Replace by \answerYes{}, \answerNo{}, or \answerNA{}.
    \item[] Justification: We provide a github repository with code, benchmark artifacts, and instructions for reproducing the main experiments, and include it as supplementary material at submission time.
    \item[] Guidelines:
    \begin{itemize}
        \item The answer \answerNA{} means that paper does not include experiments requiring code.
        \item Please see the NeurIPS code and data submission guidelines (\url{https://neurips.cc/public/guides/CodeSubmissionPolicy}) for more details.
        \item While we encourage the release of code and data, we understand that this might not be possible, so \answerNo{} is an acceptable answer. Papers cannot be rejected simply for not including code, unless this is central to the contribution (e.g., for a new open-source benchmark).
        \item The instructions should contain the exact command and environment needed to run to reproduce the results. See the NeurIPS code and data submission guidelines (\url{https://neurips.cc/public/guides/CodeSubmissionPolicy}) for more details.
        \item The authors should provide instructions on data access and preparation, including how to access the raw data, preprocessed data, intermediate data, and generated data, etc.
        \item The authors should provide scripts to reproduce all experimental results for the new proposed method and baselines. If only a subset of experiments are reproducible, they should state which ones are omitted from the script and why.
        \item At submission time, to preserve anonymity, the authors should release anonymized versions (if applicable).
        \item Providing as much information as possible in supplemental material (appended to the paper) is recommended, but including URLs to data and code is permitted.
    \end{itemize}

\item {\bf Experimental setting/details}
    \item[] Question: Does the paper specify all the training and test details (e.g., data splits, hyperparameters, how they were chosen, type of optimizer) necessary to understand the results?
    \item[] Answer: \answerYes{} % Replace by \answerYes{}, \answerNo{}, or \answerNA{}.
    \item[] Justification: Section~\ref{sec:experiment} and Appendix~\ref{bench_details} provide the data splits, model settings, hyperparameters, optimizer, and decoding details.
    \item[] Guidelines:
    \begin{itemize}
        \item The answer \answerNA{} means that the paper does not include experiments.
        \item The experimental setting should be presented in the core of the paper to a level of detail that is necessary to appreciate the results and make sense of them.
        \item The full details can be provided either with the code, in appendix, or as supplemental material.
    \end{itemize}

\item {\bf Experiment statistical significance}
    \item[] Question: Does the paper report error bars suitably and correctly defined or other appropriate information about the statistical significance of the experiments?
    \item[] Answer: \answerYes{} % Replace by \answerYes{}, \answerNo{}, or \answerNA{}.
    \item[] Justification: All reported significance results are based on a two-tailed paired test with $p<0.05$.
    \item[] Guidelines:
    \begin{itemize}
        \item The answer \answerNA{} means that the paper does not include experiments.
        \item The authors should answer \answerYes{} if the results are accompanied by error bars, confidence intervals, or statistical significance tests, at least for the experiments that support the main claims of the paper.
        \item The factors of variability that the error bars are capturing should be clearly stated (for example, train/test split, initialization, random drawing of some parameter, or overall run with given experimental conditions).
        \item The method for calculating the error bars should be explained (closed form formula, call to a library function, bootstrap, etc.)
        \item The assumptions made should be given (e.g., Normally distributed errors).
        \item It should be clear whether the error bar is the standard deviation or the standard error of the mean.
        \item It is OK to report 1-sigma error bars, but one should state it. The authors should preferably report a 2-sigma error bar than state that they have a 96\% CI, if the hypothesis of Normality of errors is not verified.
        \item For asymmetric distributions, the authors should be careful not to show in tables or figures symmetric error bars that would yield results that are out of range (e.g., negative error rates).
        \item If error bars are reported in tables or plots, the authors should explain in the text how they were calculated and reference the corresponding figures or tables in the text.
    \end{itemize}

\item {\bf Experiments compute resources}
    \item[] Question: For each experiment, does the paper provide sufficient information on the computer resources (type of compute workers, memory, time of execution) needed to reproduce the experiments?
    \item[] Answer: \answerYes{} % Replace by \answerYes{}, \answerNo{}, or \answerNA{}.
    \item[] Justification: Section~\ref{sec:experiment} and Appendix~\ref{app:training_res} report the GPU type, memory usage, and training-time statistics.
    \item[] Guidelines:
    \begin{itemize}
        \item The answer \answerNA{} means that the paper does not include experiments.
        \item The paper should indicate the type of compute workers CPU or GPU, internal cluster, or cloud provider, including relevant memory and storage.
        \item The paper should provide the amount of compute required for each of the individual experimental runs as well as estimate the total compute. 
        \item The paper should disclose whether the full research project required more compute than the experiments reported in the paper (e.g., preliminary or failed experiments that didn't make it into the paper). 
    \end{itemize}
    
\item {\bf Code of ethics}
    \item[] Question: Does the research conducted in the paper conform, in every respect, with the NeurIPS Code of Ethics \url{https://neurips.cc/public/EthicsGuidelines}?
    \item[] Answer: \answerYes{} % Replace by \answerYes{}, \answerNo{}, or \answerNA{}.
    \item[] Justification: The research uses public or synthetic data and follows the NeurIPS Code of Ethics.
    \item[] Guidelines:
    \begin{itemize}
        \item The answer \answerNA{} means that the authors have not reviewed the NeurIPS Code of Ethics.
        \item If the authors answer \answerNo, they should explain the special circumstances that require a deviation from the Code of Ethics.
        \item The authors should make sure to preserve anonymity (e.g., if there is a special consideration due to laws or regulations in their jurisdiction).
    \end{itemize}

\item {\bf Broader impacts}
    \item[] Question: Does the paper discuss both potential positive societal impacts and negative societal impacts of the work performed?
    \item[] Answer: \answerYes{} % Replace by \answerYes{}, \answerNo{}, or \answerNA{}.
    \item[] Justification: Appendix~\ref{app:broader_impact} discusses both beneficial uses and potential risks of exact copying.
    \item[] Guidelines:
    \begin{itemize}
        \item The answer \answerNA{} means that there is no societal impact of the work performed.
        \item If the authors answer \answerNA{} or \answerNo, they should explain why their work has no societal impact or why the paper does not address societal impact.
        \item Examples of negative societal impacts include potential malicious or unintended uses (e.g., disinformation, generating fake profiles, surveillance), fairness considerations (e.g., deployment of technologies that could make decisions that unfairly impact specific groups), privacy considerations, and security considerations.
        \item The conference expects that many papers will be foundational research and not tied to particular applications, let alone deployments. However, if there is a direct path to any negative applications, the authors should point it out. For example, it is legitimate to point out that an improvement in the quality of generative models could be used to generate Deepfakes for disinformation. On the other hand, it is not needed to point out that a generic algorithm for optimizing neural networks could enable people to train models that generate Deepfakes faster.
        \item The authors should consider possible harms that could arise when the technology is being used as intended and functioning correctly, harms that could arise when the technology is being used as intended but gives incorrect results, and harms following from (intentional or unintentional) misuse of the technology.
        \item If there are negative societal impacts, the authors could also discuss possible mitigation strategies (e.g., gated release of models, providing defenses in addition to attacks, mechanisms for monitoring misuse, mechanisms to monitor how a system learns from feedback over time, improving the efficiency and accessibility of ML).
    \end{itemize}
    
\item {\bf Safeguards}
    \item[] Question: Does the paper describe safeguards that have been put in place for responsible release of data or models that have a high risk for misuse (e.g., pre-trained language models, image generators, or scraped datasets)?
    \item[] Answer: \answerYes{} % Replace by \answerYes{}, \answerNo{}, or \answerNA{}.
    \item[] Justification: The random-sequence data in our work uses synthetic credential-like strings rather than real secrets.
    \item[] Guidelines:
    \begin{itemize}
        \item The answer \answerNA{} means that the paper poses no such risks.
        \item Released models that have a high risk for misuse or dual-use should be released with necessary safeguards to allow for controlled use of the model, for example by requiring that users adhere to usage guidelines or restrictions to access the model or implementing safety filters. 
        \item Datasets that have been scraped from the Internet could pose safety risks. The authors should describe how they avoided releasing unsafe images.
        \item We recognize that providing effective safeguards is challenging, and many papers do not require this, but we encourage authors to take this into account and make a best faith effort.
    \end{itemize}

\item {\bf Licenses for existing assets}
    \item[] Question: Are the creators or original owners of assets (e.g., code, data, models), used in the paper, properly credited and are the license and terms of use explicitly mentioned and properly respected?
    \item[] Answer: \answerYes{} % Replace by \answerYes{}, \answerNo{}, or \answerNA{}.
    \item[] Justification: Appendix~\ref{app:licenses} documents the external assets and their licenses or terms of use.
    \item[] Guidelines:
    \begin{itemize}
        \item The answer \answerNA{} means that the paper does not use existing assets.
        \item The authors should cite the original paper that produced the code package or dataset.
        \item The authors should state which version of the asset is used and, if possible, include a URL.
        \item The name of the license (e.g., CC-BY 4.0) should be included for each asset.
        \item For scraped data from a particular source (e.g., website), the copyright and terms of service of that source should be provided.
        \item If assets are released, the license, copyright information, and terms of use in the package should be provided. For popular datasets, \url{paperswithcode.com/datasets} has curated licenses for some datasets. Their licensing guide can help determine the license of a dataset.
        \item For existing datasets that are re-packaged, both the original license and the license of the derived asset (if it has changed) should be provided.
        \item If this information is not available online, the authors are encouraged to reach out to the asset's creators.
    \end{itemize}

\item {\bf New assets}
    \item[] Question: Are new assets introduced in the paper well documented and is the documentation provided alongside the assets?
    \item[] Answer: \answerYes{} % Replace by \answerYes{}, \answerNo{}, or \answerNA{}.
    \item[] Justification: Appendix~\ref{bench_details} documents the benchmark construction, data sources, and dataset statistics.
    \item[] Guidelines:
    \begin{itemize}
        \item The answer \answerNA{} means that the paper does not release new assets.
        \item Researchers should communicate the details of the dataset\slash code\slash model as part of their submissions via structured templates. This includes details about training, license, limitations, etc. 
        \item The paper should discuss whether and how consent was obtained from people whose asset is used.
        \item At submission time, remember to anonymize your assets (if applicable). You can either create an anonymized URL or include an anonymized zip file.
    \end{itemize}

\item {\bf Crowdsourcing and research with human subjects}
    \item[] Question: For crowdsourcing experiments and research with human subjects, does the paper include the full text of instructions given to participants and screenshots, if applicable, as well as details about compensation (if any)? 
    \item[] Answer: \answerNA{} % Replace by \answerYes{}, \answerNo{}, or \answerNA{}.
    \item[] Justification: This paper does not involve crowdsourcing or research with human subjects.
    \item[] Guidelines:
    \begin{itemize}
        \item The answer \answerNA{} means that the paper does not involve crowdsourcing nor research with human subjects.
        \item Including this information in the supplemental material is fine, but if the main contribution of the paper involves human subjects, then as much detail as possible should be included in the main paper. 
        \item According to the NeurIPS Code of Ethics, workers involved in data collection, curation, or other labor should be paid at least the minimum wage in the country of the data collector. 
    \end{itemize}

\item {\bf Institutional review board (IRB) approvals or equivalent for research with human subjects}
    \item[] Question: Does the paper describe potential risks incurred by study participants, whether such risks were disclosed to the subjects, and whether Institutional Review Board (IRB) approvals (or an equivalent approval/review based on the requirements of your country or institution) were obtained?
    \item[] Answer: \answerNA{} % Replace by \answerYes{}, \answerNo{}, or \answerNA{}.
    \item[] Justification: This paper does not involve human-subject research so IRB approval is not applicable.
    \item[] Guidelines:
    \begin{itemize}
        \item The answer \answerNA{} means that the paper does not involve crowdsourcing nor research with human subjects.
        \item Depending on the country in which research is conducted, IRB approval (or equivalent) may be required for any human subjects research. If you obtained IRB approval, you should clearly state this in the paper. 
        \item We recognize that the procedures for this may vary significantly between institutions and locations, and we expect authors to adhere to the NeurIPS Code of Ethics and the guidelines for their institution. 
        \item For initial submissions, do not include any information that would break anonymity (if applicable), such as the institution conducting the review.
    \end{itemize}

\item {\bf Declaration of LLM usage}
    \item[] Question: Does the paper describe the usage of LLMs if it is an important, original, or non-standard component of the core methods in this research? Note that if the LLM is used only for writing, editing, or formatting purposes and does \emph{not} impact the core methodology, scientific rigor, or originality of the research, declaration is not required.
    %this research? 
    \item[] Answer: \answerNA{} % Replace by \answerYes{}, \answerNo{}, or \answerNA{}.
    \item[] Justification: LLM is only used only for writing, editing, or formatting purposes.
    \item[] Guidelines:
    \begin{itemize}
        \item The answer \answerNA{} means that the core method development in this research does not involve LLMs as any important, original, or non-standard components.
        \item Please refer to our LLM policy in the NeurIPS handbook for what should or should not be described.
    \end{itemize}

\end{enumerate}

\end{document}